# A survey of dimensionality reduction techniques

C.O.S. Sorzano[1, †], J. Vargas[1], A. Pascual-Montano[1]

[1]*Natl. Centre for Biotechnology (CSIC)*

*C/Darwin, 3. Campus Univ. Autónoma, 28049 Cantoblanco, Madrid, Spain*

`{coss,jvargas,pascual}@cnb.csic.es`

[†]*Corresponding author*

*Abstract*—**Experimental life sciences like biology or chemistry have seen in the recent decades an explosion of the data available from experiments. Laboratory instruments become more and more complex and report hundreds or thousands measurements for a single experiment and therefore the statistical methods face challenging tasks when dealing with such high-dimensional data. However, much of the data is highly redundant and can be efficiently brought down to a much smaller number of variables without a significant loss of information. The mathematical procedures making possible this reduction are called dimensionality reduction techniques; they have widely been developed by fields like Statistics or Machine Learning, and are currently a hot research topic. In this review we categorize the plethora of dimension reduction techniques available and give the mathematical insight behind them.**

## 1. Introduction

During the last decade life sciences have undergone a tremendous revolution with the accelerated development of high technologies and laboratory instrumentations. A good example is the biomedical domain that has experienced a drastic advance since the advent of complete genome sequences. This post-genomics era has leaded to the development of new high-throughput techniques that are generating enormous amounts of data, which have implied the exponential growth of many biological databases. In many cases, these datasets have much more variables than observations. For example, standard microarray datasets usually are composed by thousands of variables (genes) in dozens of samples. This situation is not exclusive of biomedical research and many other scientific fields have also seen an explosion of the number of variables measured for a single experiment. This is the case of image processing, mass spectrometry, time series analysis, internet search engines, and automatic text analysis among others.

Statistical and machine reasoning methods face a formidable problem when dealing with such high-dimensional data, and normally the number of input variables is reduced before a data mining algorithm can be successfully applied. The dimensionality reduction can be made in two different ways: by only keeping the most relevant variables from the original dataset (this technique is called *feature selection*) or by exploiting the redundancy of the input data and by finding a smaller set of new variables, each being a combination of the input variables, containing basically the same information as the input variables (this technique is called *dimensionality reduction*).

This situation is not new in Statistics. In fact one of the most widely used dimensionality reduction techniques, Principal Component Analysis (PCA), dates back to Karl Pearson in 1901 [Pearson1901]. The key idea is to find a new coordinate system in which the input data can be expressed with many less variables without a significant error. This new basis can be global or local and can fulfill very different properties. The recent explosion of data available together with the evermore powerful computational resources have attracted the attention of many researchers in Statistics, Computer Science and Applied Mathematics who have developed a wide range of computational techniques dealing with the dimensionality reduction problem (for reviews see [Carreira1997, Fodor2002,Mateen2009]).

In this review we provide an up-to-date overview of the mathematical properties and foundations of the different dimensionality reduction techniques. For feature selection, the reader is referred to the reviews of [Dash1997], [Guyon2003] and [Saeys2007].

There are several dimensionality reduction techniques specifically designed for time series. These methods specifically exploit the frequential content of the signal and its usual sparseness in the frequency space. The most popular methods are those based on wavelets [Rioul1991, Graps1995], followed at a large distance by the Empirical Mode Decomposition [Huang1998, Rilling2003] (the reader is referred to the references above for further details). We do not cover these techniques here since they are not usually applied for the general purpose dimensionality reduction of data. From a general point of view, we may say that wavelets project the input time series onto a fixed dictionary (see Section 3). This dictionary has the property of making the projection sparse (only a few coefficients are sufficiently large), and the dimensionality reduction is obtained by setting most coefficients (the small ones) to zero. The empirical mode decomposition, instead, constructs a dictionary specially adapted to each input signal.

To keep the consistency of this review, we do not cover neither those dimensionality reduction techniques that take into account the class of observations, i.e., there are observations from a class A of objects, observations from a class B, … and the dimensionality reduction technique should keep as well as possible the separability of the original classes. Fisher's Linear Discriminant Analysis (LDA) was one of the first techniques to address this issue [Fisher1936, Rao1948]. Many other works followed since then, for the most recent works and for a bibliographical review see [Bian2011, Cai2011, Kim2011, Lin2011, Batmanghelich2012].

In the following we will refer to the observations as input vectors $\mathbf{x}$, whose dimension is $M$. We will assume that we have $N$ observations and we will refer to the $n$-th observation as $\mathbf{x}_n$. The whole dataset of observations will be $X$, while $X$ will be a $M \times N$ matrix with all the observations as columns. Note that small, bold letters represent vectors ($\mathbf{x}$), while capital, non-bold letters ($X$) represent matrices. The goal of the dimensionality reduction is to find another representation $\chi$ of a smaller dimension $m$ such that as much information as possible is retained from the original set of observations $\mathbf{x}$. This involves some transformation operator from the original vectors onto the new vectors, $\chi = T(\mathbf{x})$. These projected vectors are sometimes called feature vectors, and the projection of $\mathbf{x}_n$ will be noted as $\chi_n$. There might not be an inverse for this projection, but there must be a way of recovering an approximate value to the original vector, $\hat{\mathbf{x}} = R(\chi)$, such that $\hat{\mathbf{x}} \approx \mathbf{x}$.

An interesting property of any dimensionality reduction technique is to consider its stability. In this context, a technique is said to be ε-stable, if for any two input data points, $\mathbf{x}_1$ and $\mathbf{x}_2$, the following inequation holds [Baraniuk2010]: $(1-\varepsilon)\|\mathbf{x}_1 - \mathbf{x}_2\|_2^2 \leq \|\chi_1 - \chi_2\|_2^2 \leq (1+\varepsilon)\|\mathbf{x}_1 - \mathbf{x}_2\|_2^2$. Intuitively, this equation implies that Euclidean distances in the original input space are relatively conserved in the output feature space.

## 2. Methods based on Statistics and Information Theory

This family of methods reduces the input data according to some statistical or information theory criterion. Somehow, the methods based on information theory can be seen as a generalization of the ones based on statistics in the sense that they can capture non-linear relationships between variables, can handle interval and categorical variables at the same time, and many of them are invariant to monotonic transformations of the input variables.

### 2.1 Vector Quantization and Mixture models

Probably the simplest way of reducing dimensionality is by assigning a class (among a total of $K$ classes) to each one of the observations $\mathbf{x}_n$. This can be seen as an extreme case of dimensionality reduction in which we go from $M$ dimensions to 1 (the discrete class label $\chi$). Each class, $\chi$, has a representative $\overline{\mathbf{x}}_\chi$ which is the average of all the observations assigned to that class. If a vector $\mathbf{x}_n$ has been assigned to the $\chi_n$-th class, then its approximation after the dimensionality reduction is simply $\hat{\mathbf{x}}_n = \overline{\mathbf{x}}_{\chi_n}$ (see Fig. 1).

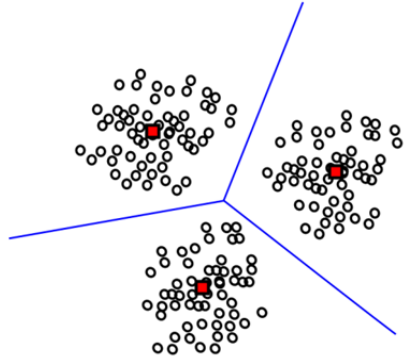

Figure 1.Example of the use of a vector quantization. Black circles represent the input data, $\mathbf{x}_n$; red squares represent class representatives, $\overline{\mathbf{x}}_\chi$.

The goal is thus to find the representatives $\overline{\mathbf{x}}_\chi$ and class assignments $u_\chi(\mathbf{x})$ ($u_\chi(\mathbf{x})$ is equal to 1 if the observation $\mathbf{x}$ is assigned to the $\chi$-th class, and is 0 otherwise) such that $J_{VQ} = E\left\{\sum_{\chi=1}^{K} u_\chi(\mathbf{x})\|\mathbf{x}-\overline{\mathbf{x}}_\chi\|^2\right\}$ is minimized. This problem is known as vector quantization or K-means [Hartigan1979]. The optimization of this goal function is a combinatorial problem although there are heuristics to cut down its cost [Gray1984,Gersho1992]. An alternative formulation of the K-means objective function is $J_{VQ} = \|X - WU\|_F^2$ subject to $U^t U = I$ and $u_{ij} \in \{0,1\}$ (i.e., that each input vector is assigned to one and only one class) [Batmanghelich2012]. In this expression, $W$ is a $M \times m$ matrix with all representatives as column vectors, $U$ is a $m \times N$ matrix whose ij-th entry is 1 if the j-th input vector is assigned to the i-th class, and $\|\cdot\|_F^2$ denotes the Frobenius norm of a matrix.

This intuitive goal function can be put in a probabilistic framework. Let us assume we have a generative model of how the data is produced. Let us assume that the observed data are noisy versions of $K$ vectors $\mathbf{x}_\chi$ which are equally likely *a priori*. Let us assume that the observation noise is normally distributed with a spherical covariance matrix $\Sigma = \sigma^2 I$. The likelihood of observing $\mathbf{x}_n$ having produced $\mathbf{x}_\chi$ is $l(\mathbf{x}_n | \mathbf{x}_\chi, \sigma^2) = \frac{1}{(2\pi)^{\frac{M}{2}} \sigma} \exp\left(-\frac{1}{2} \frac{\|\mathbf{x}_n - \mathbf{x}_\chi\|^2}{\sigma^2}\right)$. With our previous definition of $u_\chi(\mathbf{x})$ we can express it as $l(\mathbf{x}_n | \mathbf{x}_\chi, \sigma^2) = \frac{1}{(2\pi)^{\frac{M}{2}} \sigma} \exp\left(-\frac{1}{2} \frac{\sum_{\chi=1}^{K} u_\chi(\mathbf{x}_n)\|\mathbf{x}_n - \mathbf{x}_\chi\|^2}{\sigma^2}\right)$. The log likelihood of observing the whole dataset $\mathbf{x}_n$ ($n=1,2,...,N$) after removing all constants is $L(X | \mathbf{x}_\chi) = \sum_{n=1}^{N}\sum_{\chi=1}^{K} u_\chi(\mathbf{x}_n)\|\mathbf{x}_n - \mathbf{x}_\chi\|^2$. We, thus, see that the goal function of vector quantization $J_{VQ}$ produces the maximum likelihood estimates of the underlying $\mathbf{x}_\chi$ vectors.

Under this generative model, the probability density function of the observations is the convolution of a Gaussian function and a set of delta functions located at the $\mathbf{x}_\chi$ vectors, i.e., a set of Gaussians located at the $\mathbf{x}_\chi$ vectors. The vector quantization then is an attempt to find the centers of the Gaussians forming the probability density function of the input data. This idea has been further pursued by Mixture Models [Bailey1994] that are a generalization of vector quantization in which, instead of looking only for the means of the Gaussians associated to each class, we also allow each class to have a different covariance matrix $\Sigma_\chi$, and different *a priori* probability $\pi_\chi$. The algorithm looks for estimates of all these parameters by Expectation-Maximization, and at the end produces for each input observation $\mathbf{x}_n$, the label $\chi$ of the Gaussian that has the maximum likelihood of having generated that observation.

We can extend this concept and, instead of making a hard class assignment, we can make a fuzzy class assignment by allowing $0 \leq u_\chi(\mathbf{x}) \leq 1$ and requiring $\sum_{\chi=1}^{I} u_\chi(\mathbf{x}) = 1$ for all $\mathbf{x}$. This is another famous vector quantization algorithm called fuzzy K-means [Bezdek1981, Bezdek1984].

The K-means algorithm is based on a quadratic objective function, which is known to be strongly affected by outliers. This drawback can be alleviated by taking the $l_1$ norm of the approximation errors and modifying the problem to $J_{K-medians} = \|X - WU\|_1^2$ subject to $U^t U = I$ and $u_{ij} \in \{0,1\}$ [Arora1998, Batmanghelich2012]. [Iglesias2007] proposed a different approach to find data representatives less affected by outliers which we may call robust Vector Quantization, $J_{RVQ} = E\left\{\sum_{\chi=1}^{K} u_\chi(\mathbf{x}) \Phi\left(\|\mathbf{x} - \bar{\mathbf{x}}_\chi\|^2\right)\right\}$ where $\Phi(x)$ is a function less sensitive to outliers than $\Phi(x) = x$, for instance [Iglesias2007] proposes $\Phi(x) = x^\alpha$ with $\alpha$ about 0.5.

Some authors [Girolami2002, Dhillon2004, Yu2012] have proposed to use a non-linear embedding of the input vectors $\mathbf{\Phi}(\mathbf{x})$ into a higher dimensional space (dimensionality expansion, instead of reduction), and then perform the vector quantization in this higher dimensional space (this kind of algorithms are called Kernel algorithms and are further explained below with Kernel PCA). The reason for performing this non-linear mapping is that the topological spherical balls induced by the distance $\|\mathbf{\Phi}(\mathbf{x}) - \mathbf{\Phi}(\bar{\mathbf{x}}_\chi)\|^2$ in the higher-dimensional space, correspond to non-spherical neighborhoods in the original input space, thus allowing for a richer family of distance functions.

Although vector quantization has all the ingredients to be considered a dimensionality reduction (mapping from the high dimensional space to the low dimensional space by assigning a class label $\chi$, and back to high dimensional space through an approximation), this algorithm has a serious drawback. The problem is that the distances in the feature space ($\chi$ runs from 1 to $K$) do not correspond to distances in the original space. For example, if $\mathbf{x}_n$ is assigned label 0, $\mathbf{x}_{n+1}$ label 1, and $\mathbf{x}_{n+2}$ label 2, it does not mean that $\mathbf{x}_n$ is closer to $\mathbf{x}_{n+1}$ than to $\mathbf{x}_{n+2}$ in the input $M$ dimensional space. Labels are arbitrary and do not allow to conclude anything about the relative organization of the input data other than knowing that all vectors assigned to the same label are closer to the representative of that label than to the representative of any other label (this fact creates a Voronoi partition of the input space) [Gray1984, Gersho1992].

The algorithms presented from now on do not suffer from this problem. Moreover, the problem can be further attenuated by imposing a neighborhood structure on the feature space. This is done by Self-Organizing Maps and Generative Topographic Mappings, which are presented below.

## 2.2 PCA

Principal Component Analysis (PCA) is by far one of the most popular algorithms for dimensionality reduction [Pearson1901, Wold1987, Dunteman1989, Jollife2002]. Given a set of observations $\mathbf{x}$, with dimension $M$ (they lie in $\mathbb{R}^M$), PCA is the standard technique for finding the single best (in the sense of least-square error) subspace of a given dimension, $m$. Without loss of generality, we may assume the data is zero-mean and the subspace to fit is a linear subspace (passing through the origin).

This algorithm is based on the search of orthogonal directions explaining as much variance of the data as possible. In terms of dimensionality reduction it can be formulated [Hyvarinen2001] as the problem of finding the $m$ orthonormal directions $\mathbf{w}_i$ minimizing the representation error $J_{PCA} = E\left\{\left\|\mathbf{x} - \sum_{i=1}^{m} \langle \mathbf{w}_i, \mathbf{x}\rangle \mathbf{w}_i\right\|^2\right\}$. In this objective function, the reduced vectors are the projections $\boldsymbol{\chi} = (\langle \mathbf{w}_1, \mathbf{x}\rangle, ..., \langle \mathbf{w}_m, \mathbf{x}\rangle)^t$. This can be much more compactly written as $\boldsymbol{\chi} = W^t \mathbf{x}$, where $W$ is a $M \times m$ matrix whose columns are the orthonormal directions $\mathbf{w}_i$ (or equivalently $W^t W = I$). The approximation to the original vectors is given by $\hat{\mathbf{x}} = \sum_{i=1}^{m} \langle \mathbf{w}_i, \mathbf{x}\rangle \mathbf{w}_i$, or what is the same, $\hat{\mathbf{x}} = W\boldsymbol{\chi}$. In Figure 2, we show a graphical representation of a PCA transformation in only two dimensions ($\mathbf{x} \in \mathbb{R}^2$). As can be seen from Figure 2, the variance of the data in the original data space is best captured in the rotated space given by vectors $\boldsymbol{\chi} = W^t \mathbf{x}$.

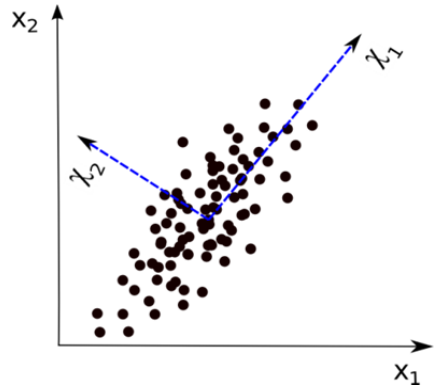

*Figure 2. Graphical representation of a PCA transformation in only two dimensions.*

$\chi_1$ is the first principal component and it goes in the direction of most variance, $\chi_2$ is the second principal component, it is orthogonal to the first and it goes in the second direction with most variance (in $\mathbb{R}^2$ there is not much choice, but in the general case, $\mathbb{R}^M$, there is). Observe that without loss of generality the data is centred about the origin of the output space.

We can rewrite the objective function as $J_{PCA} = E\left\{\|\mathbf{x} - W\boldsymbol{\chi}\|^2\right\} = E\left\{\|\mathbf{x} - WW^t \mathbf{x}\|^2\right\} \propto \|X - WW^t X\|_F^2$. Note that the class membership matrix ($U$ in vector quantization) has been substituted in this case by $W^t X$, which in general can take any positive or negative value. It, thus, has lost its membership meaning and simply constitutes the weights of the linear combination of the column vectors of $W$ that better approximate each input $\mathbf{x}$. Finally, PCA objective function can also be written as $J_{PCA} = \text{Tr}\left\{W^t \Sigma_X W\right\}$ [He2011], where $\Sigma_X = \frac{1}{N}\sum_i (\mathbf{x}_i - \bar{\mathbf{x}})(\mathbf{x}_i - \bar{\mathbf{x}})^t$ is the covariance matrix of the observed data. The PCA formulation has also been extended to complex-valued input vectors [Li2011], the method is called non-circular PCA.

The matrix projection of the input vectors onto a lower dimensional space ($\boldsymbol{\chi} = W^t \mathbf{x}$) is a wide-spread technique in dimensionality reduction as will be shown in this article. The elements involved in this projection have an interesting interpretation as explained in the following example. Let us assume that we are analyzing scientific articles related to a specific domain. Each article will be represented by a vector $\mathbf{x}$ of word frequencies, i.e., we choose a set of $M$ words representative of our scientific area, and we annotate how many times each word appears in each article. Each vector $\mathbf{x}$ is then orthogonally projected onto the new subspace defined by the vectors $\mathbf{w}_i$. Each vector $\mathbf{w}_i$ has dimension $M$ and it can be understood as a "topic" (i.e., a topic is characterized by the relative frequencies of the $M$ different words; two different topics will differ in the relative frequencies of the $M$ words). The projection of $\mathbf{x}$ onto each $\mathbf{w}_i$ gives an idea of how important is topic $\mathbf{w}_i$ to represent the article $\mathbf{x}$. Important topics have large projection values and, therefore, large values in the corresponding component of $\chi$ (see Fig. 3).

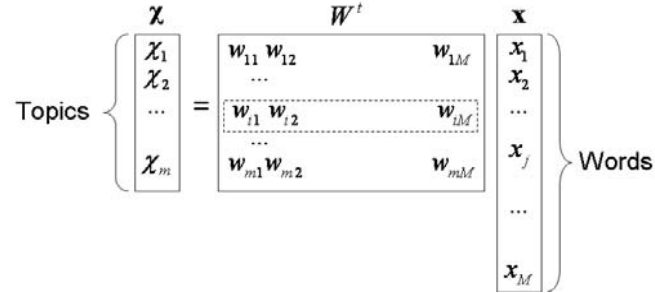

*Figure 3. Projection of the vector $\mathbf{x}$ onto the subspace spanned by the vectors $\mathbf{w}_i$ (rows of $W^t$). The components of $\mathbf{x}$ represent the frequency of each word in a given scientific article. The vectors $\mathbf{w}_i$ represent the word composition of a given topic. Each component of the projected vector $\chi$ represents how important is that topic for the article being treated.*

It can be shown [Hyvarinen2001,Jenssen2010] that when the input vectors, $\mathbf{x}$, are zero-mean (if they are not, we can transform the input data simply by subtracting the sample average vector), then the solution of the minimization of $J_{PCA}$ is given by the $m$ eigenvectors associated to the largest $m$ eigenvalues of the covariance matrix of $\mathbf{x}$ ($C_\mathbf{x} = \frac{1}{N} XX^t$, note that the covariance matrix of $\mathbf{x}$ is a $M \times M$ matrix with $M$ eigenvalues). If the eigenvalue decomposition of the input covariance matrix is $C_\mathbf{x} = W_M \Lambda_M W_M^t$ (since $C_\mathbf{x}$ is a real-symmetric matrix), then the feature vectors are constructed as $\chi = \Lambda_m^{-\frac{1}{2}} W_m^t \mathbf{x}$, where $\Lambda_m$ is a diagonal matrix with the $m$ largest eigenvalues of the matrix $\Lambda_M$ and $W_m$ are the corresponding $m$ columns from the eigenvectors matrix $W_M$. We could have constructed all the feature vectors at the same time by projecting the whole matrix $X$, $U = \Lambda_m^{-\frac{1}{2}} W_m^t X$. Note that the $i$-th feature is the projection of the input vector $\mathbf{x}$ onto the $i$-th eigenvector, $\chi_i = \lambda_i^{-\frac{1}{2}} \mathbf{w}_i^t \mathbf{x}$. The so-computed feature vectors have identity covariance matrix, $C_\chi = I$, meaning that the different features are decorrelated.

Univariate variance is a second-order statistical measure of the departure of the input observations with respect to the sample mean. A generalization of the univariate variance to multivariate variables is the trace of the input covariance matrix. By choosing the $m$ largest eigenvalues of the covariance matrix $C_\mathbf{x}$, we guarantee that we are making a representation in the feature space explaining as much variance of the input space as possible with only $m$ variables. In fact, $\mathbf{w}_1$ is the direction in which the data has the largest variability, $\mathbf{w}_2$ is the direction with largest variability once the variability along $\mathbf{w}_1$ has been removed, $\mathbf{w}_3$ is the direction with largest variability once the variability along $\mathbf{w}_1$ and $\mathbf{w}_2$ has been removed, etc. Thanks to the orthogonality of the $\mathbf{w}_i$ vectors, and the subsequent decorrelation of the feature vectors, the total variance explained by PCA decomposition can be conveniently measured as the sum of the variances of each feature, $\sigma_{PCA}^2 = \sum_{i=1}^m \lambda_i = \sum_{i=1}^m \mathrm{Var}\{\chi_i\}$.

### 2.2.1 Incremental, stream or online PCA

Let us assume that we have observed a number of input vectors **x** and we have already performed their dimensionality reduction with PCA. Let us assume that we are given new observations and we want to refine our directions $\mathbf{w}_i$ to accommodate the new vectors. In the standard PCA approach we would have to re-estimate the covariance matrix of **x** (now using all the vectors, old and new), and to recompute its eigenvalue decomposition. Incremental methods (such as Incremental PCA [Artac2002]) provide clever ways of updating our estimates of the best directions $\mathbf{w}_i$ based on the old estimates of the best directions and the new data available. In this way, we can efficiently process new data as they arrive. For this reason, incremental methods are also known as stream methods or online methods. This idea can be applied to many of the methods discussed in this review and will not be further commented.

### 2.2.2 Relationship of PCA and SVD

Another approach to the PCA problem, resulting in the same projection directions $\mathbf{w}_i$ and feature vectors $\chi$ uses Singular Value Decomposition (SVD, [Golub1970, Klema1980, Wall2003]) for the calculations. Let us consider the matrix $X$ whose columns are the different observations of the vector **x**. SVD decomposes this matrix as the product of three other matrices $X = WDU$ ($W$ is of size $M \times M$, $D$ is a diagonal matrix of size $M \times N$, and $U$ is of size $N \times N$) [Abdi2007]. The columns of $W$ are the eigenvectors of the covariance matrix of **x** and $D_{ii}$ is the square root of its associated eigenvalue. The columns of $U$ are the feature vectors. So far we have not performed any dimensionality reduction yet. As in PCA the dimensionality reduction is achieved by discarding those components associated to the lowest eigenvalues. If we keep the $m$ directions with largest singular values, we are approximating the data matrix by $\hat{X} = W_m D_m U_m$ ($W_m$ is of size $M \times m$, $D_m$ is of size $m \times m$, and $U_m$ is of size $m \times N$). It has been shown [Johnson1963] that $\hat{X}$ is the matrix better approximating $X$ in the Frobenius norm sense (i.e., exactly the same as required by $J_{PCA}$).

Another interesting property highlighted by this decomposition is that the eigenvectors $\mathbf{w}_i$ (columns of the $W$ matrix) are an orthonormal basis of the subspace spanned by the observations **x**. That means that for each element in this basis we can find a linear combination of input vectors such that $\mathbf{w}_i = \sum_{n=1}^{N} \alpha_{in} \mathbf{x}_n$. This fact will be further exploited by Sparse PCA and Kernel PCA. This digression on the SVD approach to PCA helps us to understand a common situation in some experimental settings. For instance, in microarray experiments, we have about 50 samples and 1000 variables. As shown by the SVD decomposition, the rank of the covariance matrix is the minimum between $M = 1000$ and $N = 50$, therefore, we cannot compute more than 50 principal components.

An interesting remark on the SVD decomposition is that among all possible matrix decompositions of the form $X = WDU$, SVD is the only family of decompositions (SVD decomposition is not unique) yielding diagonal matrices in $D$. In other words, the matrices $W$ and $U$ can differ significantly from the SVD decomposition as long as $D$ is not a diagonal matrix. This fact is further exploited by Sparse Tensor SVD (see dictionary-based methods below).

### 2.2.3 Nonlinear PCA

PCA can be extended to non-linear projections very easily conceptually although its implementation is more involved. Projections can be replaced by $\mathbf{f}(W^t \mathbf{x})$, being $\mathbf{f}(\chi): \mathbb{R}^m \to \mathbb{R}^m$ a non-linear function chosen by the user. The goal function is then $J_{NLPCA} = E\left\{\left\|\mathbf{x} - W\mathbf{f}(W^t \mathbf{x})\right\|^2\right\}$ which is minimized subject to $W^t W = I$ [Girolami1997b].

### 2.2.4 PCA rotations and Sparse PCA

A drawback of PCA is that the eigenvectors ($\mathbf{w}_i$) have usually contributions from all input variables (all their components are significantly different from zero). This makes their interpretation more difficult since a large feature value cannot be attributed to a few (ideally a single) input values. Possible solutions are rotating the eigenvectors using PCA rotation methods (Varimax, Quartimax, etc.) or forcing many of the components of $\mathbf{w}_i$ to be zero, i.e., to have a Sparse PCA.

An easy way of forcing the interpretability of the principal components $\mathbf{w}_i$ is by rotating them once they have been computed. The subspace spanned by the rotated vectors is the same as the one spanned by the unrotated vectors. Therefore, the approximation $\hat{\mathbf{x}}$ is still the same, although the feature vector must be modified to account for the rotation. This is the approach followed by Varimax [Kaiser958]. It looks for the rotation that maximizes the variance of the feature vector after rotation (the idea is that maximizing this variance implies that each feature vector uses only a few eigenvectors). Quartimax looks for a rotation minimizing the number of factors different from zero in the feature vector. Equimax and Parsimax are compromises between Varimax and Quartimax. All these criteria are generally known as Orthomax and they have been unified under a single rotation criterion [Crawford1970]. Among these criteria, Varimax is by far the most popular. If the orthogonality condition of the vectors $\mathbf{w}_i$ is removed, then the new principal components are free to move and the rotation matrix is called an "oblique" rotation matrix. Promax [Abdi2003] is an example of such an oblique rotation technique. However, these oblique rotations have been superseded by Generalized PCA.

Recently, there have been some papers imposing the sparseness of the feature vectors by directly regularizing a functional solving the PCA problem (formulated as a regression problem). This kind of methods is commonly called Sparse PCA. One of the most popular algorithms of this kind is the one of Zou [Zou2006]. We have already seen that the PCA problem can be seen as regression problem whose objective function is $J_{PCA} = E\left\{\left\|\mathbf{x} - WW^t\mathbf{x}\right\|^2\right\}$. Normally the optimization is performed with the constraint $W^tW = I$ (i.e., the directions $\mathbf{w}_i$ have unit module). We can generalize this problem, and instead of using the same matrix to build the feature vectors ($W^t$) and reconstruct the original samples ($W$), we can make them different. We can use ridge regression (a Tikhonov regularization to avoid possible instabilities caused by the eventual ill-conditioning of the regression, the most common one is simply the $l_2$ norm), and a regularization based on some norm promoting sparseness (like the $l_1$ norm): $J_{SPCA} = E\left\{\left\|\mathbf{x} - W\tilde{W}^t\mathbf{x}\right\|^2\right\} + \lambda_1 \left\|\tilde{W}\right\|_{l_1} + \lambda_2 \left\|\tilde{W}\right\|_{l_2}^2$. In the previous formula the $l_p$ norms of the matrices are computed as $\left\|W\right\|_{l_p} = \left(\sum_{i,j}\left|w_{ij}\right|^p\right)^{\frac{1}{p}}$ and the objective function is optimized with respect to $W$ and $\tilde{W}$ that are $M \times m$ matrices. It has been shown [Zou2006] that promoting the sparseness of $\tilde{W}$ promotes the sparseness of the feature vectors which is, after all, the final goal of this algorithm.

The previous sparse approaches tried to find sparse projection directions by zeroing some of the elements in the projection directions. However, we may prefer absolutely removing some of the input features (so that their contribution to all projection directions is zero). [Ulfarsson2011] proposed the sparse variable PCA (svPCA). svPCA is based on noisy PCA (nPCA), which is a special case of Factor Analysis (see below). The observed data is supposed to have been generated as $\mathbf{x} = W\chi + \mathbf{n}$. Assuming that the covariance of the noise is $\sigma^2 I$, the goal of nPCA is to maximize the log-likelihood of observing the data matrix $X$ under the nPCA model, that is $J_{nPCA} = -\frac{1}{2}\text{Tr}\left\{\Sigma_X \Omega^{-1}\right\} - \frac{1}{2}\log|\Omega|$ where $\Omega = WW^t + \sigma^2 I$. svPCA objective function is $J_{svPCA} = J_{nPCA} - \frac{N}{2\sigma^2}\sum_{i=1}^{m}\left\|\mathbf{w}_i\right\|_0$.

### 2.2.5 Localized PCA and Subspace segmentation

As we have explained above, given a set of data samples in $\mathbb{R}^M$, the standard technique for finding the single best (in the sense of least-square error) subspace of a given dimension is the Principal Component Analysis (PCA). However, in practice, a single linear model has severe limitations for modelling real data. Because the data can be heterogeneous and contain multiple subsets each of which is best fitted with a different linear model. Suppose that we have a high dimensional dataset, that can be conveniently decomposed into different small datasets or clusters, which can be approximated well by different linear subspaces of small dimension by means of a principal component analysis. This situation is presented in Fig. 4. Note from Fig. 4 that the dataset composed by the red and blue points can be decomposed in two sets, one formed by the red and the other by the blue ones and each of this groups can effectively be approximated using a two dimensional linear subspaces.

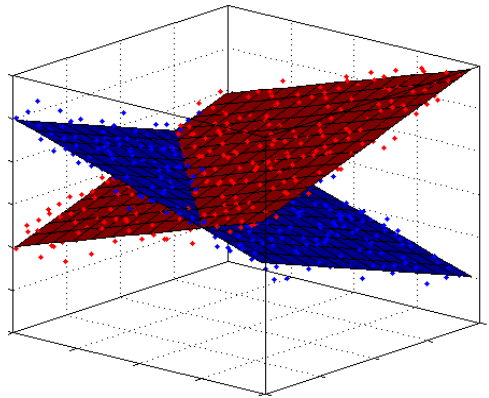

*Figure 4. Mixed dataset composed by the red and blue points that can be decomposed in two smaller datasets, which can effectively be described using two dimensional linear subspaces.*

The term Localized PCA has been used several times through literature to refer to different algorithms. Here we will refer to the most successful ones. [Fukunaga1971] proposed an extension of the K-means algorithm which we will refer to as Cluster-PCA. In K-means, a cluster is represented by its centroid. In Cluster-PCA, a cluster is represented by a centroid plus an orthogonal basis defining a subspace that embeds locally the cluster. An observation $\mathbf{x}$ is assigned to a cluster if the projection of $\mathbf{x}$ onto the cluster subspace ($\hat{\mathbf{x}} = WW^t\mathbf{x}$) is the closest one (the selection of the closest subspace must be done with care so that extrapolation of the cluster is avoided). Once all observations have been assigned to their corresponding clusters, the cluster centroid is updated as in K-means and the cluster subspace is recalculated by using PCA on the observations belonging to the cluster. As with K-means, a severe drawback of the algorithm is its dependence with the initialization, and several hierarchically divisive algorithms have been provided (Recursive Local PCA) [Liu2003b]. For a review on this kind of algorithms see [Einbeck2008].

Subspace segmentation extends the idea of locally embedding the input points into linear subspaces. The assumption is that the data has been generated using several subspaces that may not be orthogonal. The goal is to identify all these subspaces. Generalized PCA [Vidal2005] is a representative of this family of algorithms. Interestingly, the subspaces to be identified are represented as polynomials whose degree is the amount of subspaces to identify and whose derivatives at a data point give normal vectors to the subspace passing through the point.

## 2.3 Principal curves, surfaces and manifolds

PCA is the perfect tool to reduce data that in their original $M$-dimensional space lie in some linear manifold. However, there are situations at which the data follow some curved structure (e.g., a slightly bent line). In this case, approximating the curve by a straight line will not perform a good approximation of the original data. We can observe a situation of this type in Fig. 5.

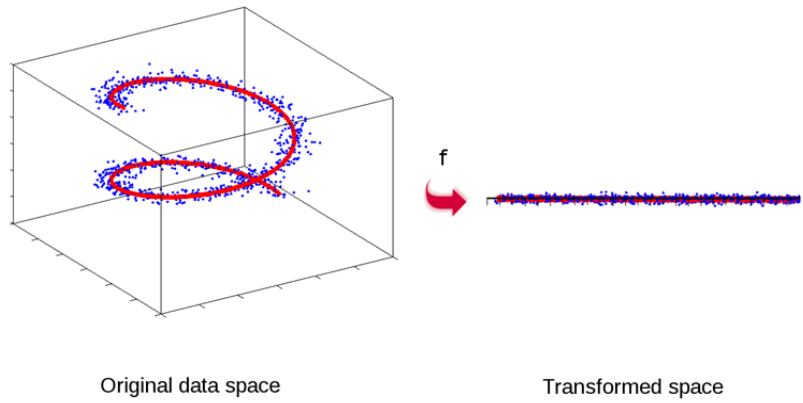

Original data space          Transformed space

*Figure 5. Dataset that lie in a curved structure (left) and transformed dataset (right)*

In Fig. 5 we show a dataset following a curved structured and therefore this dataset will not be conveniently described using the PCA method. Note that in the case of the data shown in Fig. 5, it will be needed at least three principal components to describe the data precisely. In Fig. 5 we show the same dataset after transforming it. Note that this data does no longer follow a curved a structured and in this case, it follows a linear one. Therefore, the data shown on the right of Fig. 5 can be conveniently described using PCA approach and using only one principal component.

Before introducing principal curves, surfaces and manifolds in depth, let us review PCA from a different perspective. Given a set of observations of the input vectors $\mathbf{x}$ with zero average (if the original data is not zero-average, we can simply subtract the average from all data points), we can look for the line passing through the origin and with direction $\mathbf{w}_1$ (whose equation is $\mathbf{f}(\chi) = \mathbf{w}_1 \chi$) that better fits this dataset, i.e., that minimizes $J_{line} = E\left\{ \inf_{\chi} \|\mathbf{x} - \mathbf{f}(\chi)\|^2 \right\}$. The infimum in the previous objective function implies that for each observation $\mathbf{x}_n$ we have to look for the point in the line (defined by its parameter $\chi_n$) that is closest to it. The point $\mathbf{f}(\chi_n)$ is the orthogonal projection of the observation onto the line. It can be proved that the solution of this problem is the direction with the largest data variance, that is, the same solution as in PCA. Once we have found the first principal line, we can look for the second simply by constructing a new dataset in which we have subtracted the line previously computed ($\mathbf{x}'_n = \mathbf{x}_n - \mathbf{f}(\chi_n)$). Then, we apply the same procedure $m-1$ times to detect the subsequent most important principal lines. The dimensionality reduction is achieved simply by substituting the observation $\mathbf{x}_n$ by the collection of parameters $\chi_n$ needed for its projection onto the different principal lines.

The objective function $J_{line}$ is merely a linear regression on the input data. For detecting curves instead of lines, one possibility would be to fix the family of curves sought (parabolic, hyperbolic, exponential, polynomial ...) and optimize its parameters (as was done with the line). This is exactly non-linear regression and several methods based on Neural Networks (as sophisticated non-linear regressors) have been proposed (Non-linear PCA, [Kramer1991, Scholz2008], autoencoder neural networks [Baldi1989,DeMers1993,Kambhatla1997]).

Alternatively, we can look for the best curve (not in a parametric family) [Hastie1989]. In Statistics, it is well known that the best regression function is given by the conditional expectation $\mathbf{f}(\chi) = E\{\mathbf{x} | \chi_{\mathbf{f}}(\mathbf{x})\}$, where $\chi_{\mathbf{f}}(\mathbf{x})$ represents the curve parameter $\chi$ needed to project $\mathbf{x}$ onto $\mathbf{f}$. In other words, the best curve is the one that assigns for each $\chi$ the average of all observed values projected onto $\chi$. The parameterization of the curve $\mathbf{f}$ must be such that it has unit speed (i.e., $\left\|\frac{d\mathbf{f}}{d\chi}\right\| = 1$ for all $\chi$), otherwise we could not uniquely determine this function. There are two warnings on this approach. The first one is that it might be locally biased if the noise of the observations is larger than the local curvature of the function. The second one is that if we only have a finite set of observations $\mathbf{x}$, we will have to use some approximation of the expectation so that we make the curve continuous. The two most common choices to make the curve continuous are kernel estimates of the expectation and the use of splines. In fact, the goal function of the classical smoothing spline is $J_{spline} = E\left\{\inf_{\chi}\|\mathbf{x}-\mathbf{f}(\chi)\|^2\right\} + \lambda \int \left\|\frac{d\mathbf{f}(\chi)}{d\chi}\right\|^2 d\chi$, which regularizes the curve fitting problem with the curvature of the curve. An advantage of the use of splines is their efficiency (the algorithm runs as $O(N)$ as compared to the $O(N^2)$ of the kernel estimates). However, it is difficult to choose the regularization weight, $\lambda$.

Principal Curves can be combined with the idea of Localized PCA (constructing local approximations to data). This has been done by several authors: Principal Curves of Oriented Points (PCOP) [Delicado2001], and Local Principal Curves (LPC) [Einbeck2005].

The Principal Curves idea can be extended to more dimensions (see Fig. 6). Principal surfaces are the functions minimizing $J_{surface} = E\left\{\inf_{\chi_1,\chi_2}\|\mathbf{x}-\mathbf{f}(\chi_1,\chi_2)\|^2\right\}$. The solution is again, $\mathbf{f}(\chi_1,\chi_2) = E\{\mathbf{x}\,|\,\chi_{\mathbf{f}}(\mathbf{x})\}$, where $\chi_{\mathbf{f}}(\mathbf{x})$ returns the parameters of the surface needed for the projection of $\mathbf{x}$. Intuitively, the principal surface at the point $(\chi_1,\chi_2)$ is the average of all observations whose orthogonal projection is at $\mathbf{f}(\chi_1,\chi_2)$. The extension to manifolds is straightforward [Smola1999], $J_{manifold} = E\left\{\inf_{\chi}\|\mathbf{x}-\mathbf{f}(\chi)\|^2\right\} + \lambda \|P\mathbf{f}\|^2$ that is a Tikhonov regularized version of the non-linear regression problem. $P$ is a homogenous invariant scalar operator penalizing unsmooth functions. The fact that the regularization is homogeneous invariant implies that all surfaces which can be transformed into each other by rotations are equally penalized. A feasible way of defining the function $\mathbf{f}(\chi)$ is by choosing a number of locations $\chi_i$ (normally distributed on a regular grid although the method is not restricted to this choice) and expanding this function as a weighted sum of a kernel function $k(\chi-\chi_i)$ at those locations, $\mathbf{f}(\chi) = \sum_{i=1}^{K} \mathbf{\alpha}_i k(\chi-\chi_i)$. The number of locations, $K$, controls the complexity of the manifold and the vectors $\mathbf{\alpha}_i$ (which are in the space of $\mathbf{x}$ and, thus, have dimension $M$) control its shape. However, the dimensionality reduction is still controlled by the dimension of the vector $\chi$. Radial basis functions such as the Gaussian are common kernels ($k(\chi,\chi_i) = k(\|\chi-\chi_i\|)$). With this expansion, the regularization term becomes [Smola1999] $\|P\mathbf{f}\|^2 = \sum_{i,j=1}^{I} \langle \mathbf{\alpha}_i, \mathbf{\alpha}_j \rangle k(\chi_i,\chi_j)$. An interesting feature of this approach is that by using periodical kernels, one can learn circular manifolds.

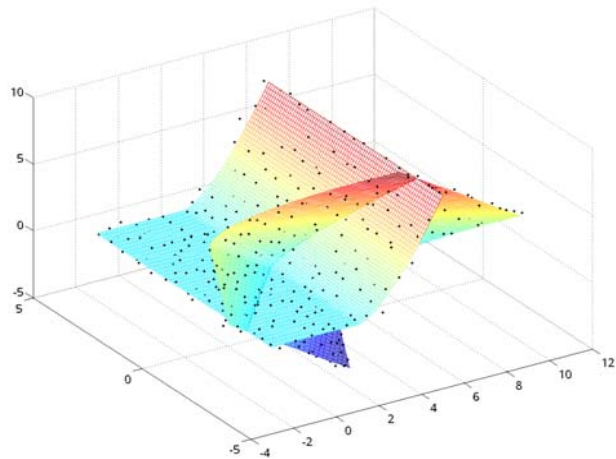

*Figure 6. Example of data distributed along two principal surfaces.*

## 2.4 Generative Topographic Mapping

A different but related approach to reduce the dimensionality by learning a manifold is the Generative Topographic Mapping (GTM) [Bishop1998]. This method is "generative" because it assumes that the observations $\mathbf{x}$ have been generated by noisy observations of a mapping of the low dimensional vectors $\chi$ onto a higher dimension, $\mathbf{f}(\chi)$ (see Fig. 7, in fact the form of this non-linear mapping is exactly the same as in the principal manifolds of the previous section, $\mathbf{f}(\chi) = \sum_{i=1}^{K} \boldsymbol{\alpha}_i k(\chi - \chi_i)$ ). In this method, it is presumed that the possible $\chi$ vectors lie on a discrete grid with $K$ points, and that the *a priori* probability of each one of the points of the grid is the same (uniform distribution). If the noise is supposed to be Gaussian (or any other spherical distribution), the maximum likelihood estimates of the vectors $\boldsymbol{\alpha}_i$ boils down to the minimization of $J_{GTM} = E\left\{\inf_{\chi} \|\mathbf{x} - \mathbf{f}(\chi)\|^2\right\}$. This objective function can be regularized by $\sum_{i=1}^{K} \|\boldsymbol{\alpha}_i\|^2$ (instead of $\|P\mathbf{f}\|^2$) which is the result of estimating the Maximum *a Posteriori* under the assumption that the $\boldsymbol{\alpha}_i$ are normally distributed with 0 mean.

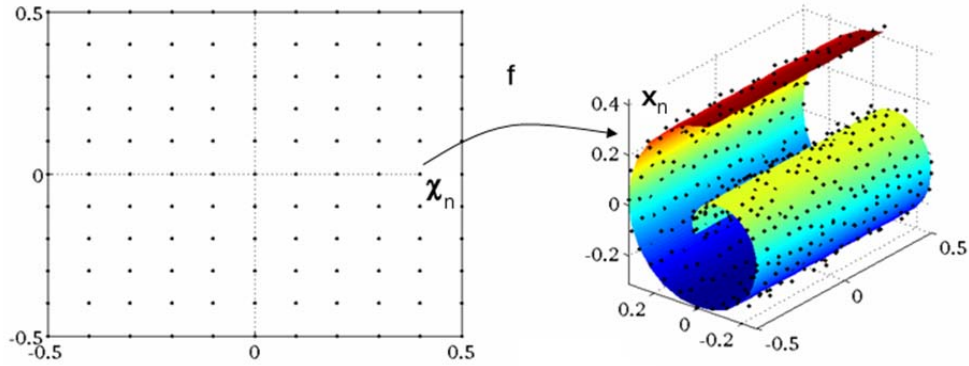

*Figure 7. Example of Generative Topographic Mapping. The observed data (right) is assumed to be generated by mapping points in a lower dimensional space.*

## 2.5 Self-Organizing Maps

In fact, GTM has been proposed as a generalization of Self-Organizing Maps, which in their turn are generalizations of the vector quantization approaches presented at the beginning. Self-Organizing Maps (SOM) work as in Vector Quantization by assigning to each input vector a label $\chi_n$ corresponding to the class closest to its representative vector. The reconstruction of $\mathbf{x}_n$ is still $\hat{\mathbf{x}}_n = \overline{\mathbf{x}}_{\chi_n}$, i.e., the class representative of class $\chi_n$. However, class labels are forced to lie in a manifold at which a topological neighborhood is defined (see Fig. 8). In this way, classes that are close to each other in the feature space are also close to each other in the input space.

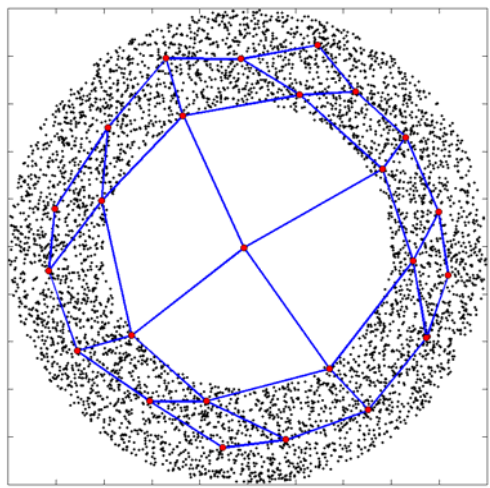

*Figure 8. Original data lies in a ring, vector representatives calculated by SOM are represented as red points. The output map topology has been represented by linking each representative vector to its neighbors with a blue edge. Note that because of the topological constraint there might be representative vectors that are not actually representing any input point (e.g., the vector in the center of the ring).*

Kohonen's SOMs [Kohonen1990, Kohonen1993, Kohonen2001] are the most famous SOMs. They work pretty well in most contexts, they are very simple to understand and implement, but they lack from a solid mathematical framework (they are not optimizing any functional and they cannot be put in a statistical framework). They start by creating a set of labels on a given manifold (usually a plane). Labels are distributed in a regular grid and the topological neighbourhood is defined as the neighbours in the plane of each point of the grid (note that this idea can be easily extended to higher dimensions). For initialization we assign to each label a class representative at random. Each input observation $\mathbf{x}_n$ is compared to all class representatives and it is assigned to the closest class whose label we will refer to as $\chi_n$. In its batch version, once all the observations have been assigned, the class representatives are then updated according to $\overline{\mathbf{x}}_\chi = \frac{\sum_{n=1}^{N} k(\chi, \chi_n) \mathbf{x}_n}{\sum_{n=1}^{N} k(\chi, \chi_n)}$. The function $k(\chi, \chi_n)$ is a kernel that gives more weight to pairs of classes that are closer in the manifold. The effect of this is that when an input vector $\mathbf{x}_n$ is assigned to a given class, the classes surrounding this class will also be updated with that input vector (although with less weight than the winning class). Classes far from the winning class receive updates with a weight very close to 0. This process is iterated until convergence.

GTM generalizes Kohonen's SOM because the class representatives $\overline{\mathbf{x}}_\chi$ in SOMs can be assimilated to the $\mathbf{f}(\chi_i)$ elements of GTM, and the function $\mathbf{f}(\chi)$ of GTM can be directly be computed from the kernel $k(\chi, \chi_n)$ in SOM [Bishop1998]. However, GTM has the advantage over SOMs that they are clearly defined in a statistical framework and the function $\mathbf{f}(\chi)$ is maximizing the likelihood of observing the given data. There is another difference of practical consequences, while SOM makes the dimensionality reduction by assigning one of the points of the grid in the manifold (i.e., it produces a discrete dimensionality reduction), GTM is capable of producing a continuous dimensionality reduction by choosing the parameters $\chi_n$ such that $\|\mathbf{x}_n - \mathbf{f}(\chi_n)\|$ is minimized.

Other generalizations of SOMs in the same direction are the Fuzzy SOM [Pascual-Marqui2001] and the Kernel Density SOM (KenDerSOM) [Pascual-Montano2001]. These generalizations rely on the regularization of the objective functions of Vector Quantization and the Mixture Models, respectively, by the term $\sum_{\chi, \chi'=1}^{K} \|\overline{\mathbf{x}}_\chi - \overline{\mathbf{x}}_{\chi'}\|^2 k(\chi, \chi')$, that is, class representatives corresponding to labels close to each other in the manifold

should have smaller differences. For a review on SOM and its relationships to Non-Linear PCA and Manifold learning see [Yin2008].

Neural Gas networks [Martinetz1991, Martinetz1993, Fritzke1995] is an approach similar to the standard SOM, only that the neighborhood topology is automatically learnt from the data. Edges appear and disappear following an aging strategy. This automatic topology learning allows adapting to complex manifolds with locally different intrinsic dimensionality [Pettis1979, Kegl2002, Costa2004] (see Fig. 9).

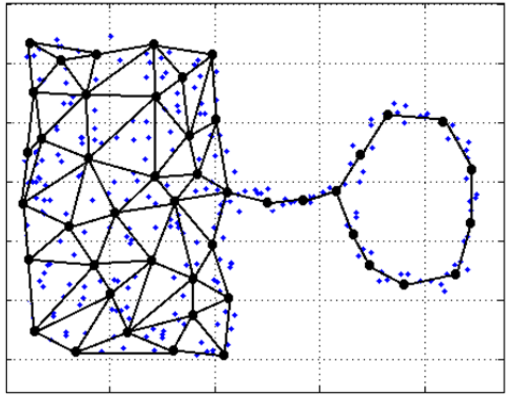

*Figure 9. Example of Neural Gas network. Note that the network has been able to learn the 2D topology present at the left points, and the 1D topology of the right points.*

## 2.6 Elastic maps, nets, principal graphs and principal trees

Elastic maps and nets [Gorban2004, Gorban2007] are half-way between SOMs and GTM. Like these two methods, the elastic map can be thought of as a net of nodes in a low-dimensional space. For each node, there is a mapping between the low-dimensional node and the high-dimensional space (like in SOM), i.e., for each node in the net $\chi$ there is a corresponding vector in the input space $\overline{\mathbf{x}}_\chi$. If an input vector $\mathbf{x}_n$ is assigned to a node $\chi_n$, then its representative is $\hat{\mathbf{x}}_n = \overline{\mathbf{x}}_{\chi_n}$ (as in SOM). The goal function combines similarity to the input data with regularity and similarity within the net:

$$J_{EN} = \sum_{n=1}^{N}\left\|\mathbf{x}_n - \overline{\mathbf{x}}_{\chi_n}\right\|^2 + \sum_{\chi,\chi'=1}^{K}\left(\lambda_\chi \left\|\overline{\mathbf{x}}_\chi - \overline{\mathbf{x}}_{\chi'}\right\|^2 g(\chi,\chi')\right) + \sum_{\chi}^{K} \mu_\chi \left\|\overline{\mathbf{x}}_\chi - \frac{\sum_{\chi'=1}^{K} g(\chi,\chi')\overline{\mathbf{x}}_{\chi'}}{\sum_{\chi'=1}^{K} g(\chi,\chi')}\right\|^2$$

. The first term accounts for the fidelity of the data representation. In the second and third terms, $g(\chi,\chi')$ define a neighborhood (is equal to 1 if the two nodes are neighbors, and equal to 0 otherwise). The second term reinforces smoothness of the manifold by favoring similarity between neighbors; the third term imposes smoothness in a different way: a node representative must be close to the average of its neighbors. As opposed to SOM, elastic nets can delete or add nodes adaptively, creating nets that are well adapted to the manifold topology. For this reason, this method is also known as Principal Graphs. Interestingly, the rules to create and delete nodes, can force the graph to be a tree.

## 2.7 Kernel PCA and Multidimensional Scaling

Kernel PCA [Scholkopf1997, Scholkopf1999] is another approach trying to capture non-linear relationships. It can be well understood if Multidimensional Scaling (MDS), another linear method, is presented first. We have already seen that PCA can be computed from the eigenvalue decomposition of the input covariance matrix, $C_\mathbf{x} = \frac{1}{N} XX^t$. However, we could have built the inner product matrix (Gram matrix) $G_\mathbf{x} = X^t X$, i.e., the $i,j$-th component of this matrix has the inner product of $\mathbf{x}_i$ with $\mathbf{x}_j$. The eigendecomposition of the Gram matrix yields $G_\mathbf{x} = W_N \Lambda_N W_N^t$ (since $G_\mathbf{x}$ is a real, symmetric matrix). MDS is a classical statistical technique [Kruskal1964a, Kruskal1964b, Schiffman1981, Kruskal1986, Cox2000, Borg2005] that builds with the $m$ largest eigenvalues a feature matrix given by $U = \Lambda^{-\frac{1}{2}} W^t$. This feature matrix is the one best preserving the inner products of the input vectors, i.e., it minimizes the Frobenius norm of the difference $G_\mathbf{x} - G_\chi$. It can be proven [Jenssen2010] that the eigenvalues of $G_\mathbf{x}$ and $C_\mathbf{x}$ are the same, that $\eta = \text{rank}(G_\mathbf{x}) = \text{rank}(C_\mathbf{x}) \leq M$, and that for any $m \leq \eta$, the space spanned by MDS and the space spanned by PCA are identical, that means that one could find a rotation such that $U_{MDS} = U_{PCA}$. Additionally, MDS can also be computed even if the data matrix $X$ is unknown, all we need is the Gram matrix, or alternatively a distance matrix. This is a situation rather common in some kind of data analysis [Cox2000].

Kernel PCA [Scholkopf1997, Scholkopf1999] is another approach trying to capture non-linear relationships. It uses a function $\boldsymbol{\Phi}$ transforming the input vector $\mathbf{x}$ onto a new vector $\boldsymbol{\Phi}(\mathbf{x})$ whose dimension is larger than $M$ (a kind of dimensionality "expansion"). However, if we choose $\boldsymbol{\Phi}$ well enough, the data in this new space may become more linear (e.g., following a straight line instead of a curve). In this new space we can perform the standard PCA and obtain the feature vector $\chi$. In Fig. 10, we show an example of the use of this multidimensional reduction method. In Fig. 10(a) it is shown a dataset (red circles) following a curved structured that cannot be described conveniently using linear PCA as the black dashed line does not describe conveniently the dataset. Therefore, to correctly describing this dataset by the standard PCA method we will need at least two principal components. In Fig. 10(b) we show the dataset after transforming it by $\boldsymbol{\Phi}$. As can be seen in this transformed and expanded space we can describe accurately the dataset using only one principal component as the dataset follows a linear relationship.

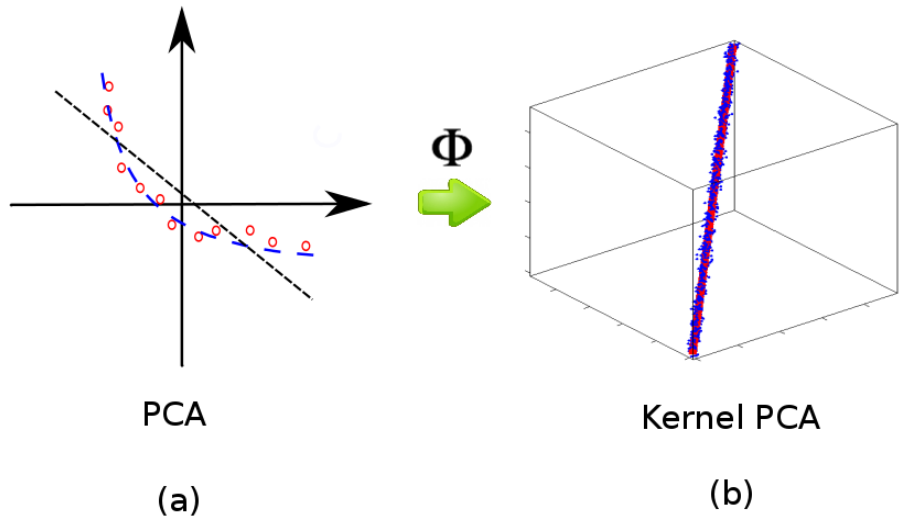

*Figure 10. Input dataset (red circles) lying in a curved structure and its corresponding first principal component using standard PCA method (black dashed line) (a). Transformed dataset by function $\boldsymbol{\Phi}$ (blue points) and its corresponding first principal component of the expanded dataset (red line) (b)*

Making use of the relationship between MDS and PCA, we do not need to compute the covariance of the $\boldsymbol{\Phi}$ vectors, but we can compute their inner products instead. We will do so through a Mercer kernel which defines a valid inner product in the space of $\boldsymbol{\Phi}$ making use of the input vectors, $\langle \boldsymbol{\Phi}(\mathbf{x}), \boldsymbol{\Phi}(\mathbf{y}) \rangle = k(\mathbf{x}, \mathbf{y})$. Common kernels are $k(\mathbf{x},\mathbf{y}) = \langle \mathbf{x},\mathbf{y}\rangle^\beta$, $k(\mathbf{x},\mathbf{y}) = \exp\left(-\frac{1}{2}\left\|\frac{\mathbf{x}-\mathbf{y}}{\beta}\right\|^2\right)$, and $k(\mathbf{x},\mathbf{y}) = \tanh\left(\beta_1\langle \mathbf{x},\mathbf{y}\rangle + \beta_2\right)$ (where the $\beta$ parameters define the kernel shape). PCA vectors $\mathbf{w}_i$ are the eigenvectors of the covariance matrix of the transformed vectors $\mathbf{\Phi}(\mathbf{x})$; but these vectors are never explicitly built neither their covariance matrix. Instead, the orthogonal directions $\mathbf{w}_i$ are computed as a linear combination of the observed data, $\mathbf{w}_i = \sum_{n=1}^{N} \alpha_{in}\mathbf{x}_n = X\boldsymbol{\alpha}_i$. The $\boldsymbol{\alpha}_i$ vectors are computed as the eigenvectors of a matrix $G_\Phi$ whose $ij$-th entry is the inner product $\langle \mathbf{\Phi}(\mathbf{x}_i),\mathbf{\Phi}(\mathbf{x}_j)\rangle$. The feature vectors can finally be computed as $\chi_i = \sum_{n=1}^{N}\alpha_{in}\langle \mathbf{\Phi}(\mathbf{x}_n),\mathbf{\Phi}(\mathbf{x})\rangle$. Obtaining the approximation of the original vector $\hat{\mathbf{x}}$ is not as straight-forward. In practice it is done by looking for a vector $\hat{\mathbf{x}}$ that minimizes $\|\mathbf{\Phi}(\mathbf{x}) - \mathbf{\Phi}(\hat{\mathbf{x}})\|^2$. The minimization is performed numerically starting from a solution specifically derived for each kernel. Again thanks to the kernel "magic", only the dot product of the transformed vectors are needed during the minimization.

All these techniques together with Locally Linear Embedding and ISOMAP (see below) are called spectral dimensionality reduction techniques because they are based on the eigenvalue decomposition of some matrix. [Bengio2006] provides an excellent review of them.

## 2.8 Kernel Entropy Component Analysis

We have already seen that PCA looks for directions that maximize the input variance explained by the feature vectors. Variance is a statistical second order measurement reflecting the amount of information contained by some variables (if there is no variability of the input vectors, there is no information in them). However, variance is a limited measure of information. Renyi's quadratic entropy is a more appropriate measure of the input information. Renyi's quadratic entropy is defined as $H(\mathbf{x}) = -\log E\{p(\mathbf{x})\}$, where $p(\mathbf{x})$ is the multivariate probability density function of the input vectors $\mathbf{x}$. Since the logarithm is a monotonic function, we may concentrate on its argument: the expectation of $p(\mathbf{x})$. However, the true underlying probability density function is unknown. We can approximate it through a kernel estimator $\hat{p}(\mathbf{x}) = \frac{1}{N}\sum_{n=1}^{N} k(\mathbf{x},\mathbf{x}_n)$, where $k(\mathbf{x},\mathbf{y})$ is a Parzen window (it is also required to be a Mercer kernel). If we now estimate Renyi's quadratic entropy as the sample average of the Parzen estimator, we get $E\{p(\mathbf{x})\} \approx \frac{1}{N}\sum_{n=1}^{N}\hat{p}(\mathbf{x}_n) = \frac{1}{N^2}\sum_{n_1=1}^{N}\sum_{n_2=1}^{N}k(\mathbf{x}_{n_1},\mathbf{x}_{n_2})$, which in the light of our previous discussion on Kernel PCA can be rewritten in terms of the Gram matrix of the $\mathbf{\Phi}$ vectors $E\{p(\mathbf{x})\} \approx \frac{1}{N^2}\mathbf{1}^t G_\Phi \mathbf{1}$ (being $\mathbf{1}$ a vector of ones with dimension $N$). The eigendecomposition of the Gram matrix yields $E\{p(\mathbf{x})\} \approx \frac{1}{N^2}\mathbf{1}^t W_N \Lambda_N W_N^t \mathbf{1} = \frac{1}{N^2}\sum_{n=1}^{N}\lambda_n \langle \mathbf{1},\mathbf{w}_n\rangle^2$, this means that for maximizing the information carried by the feature vectors is not enough choosing the eigenvectors with the $m$ largest eigenvalues, but the eigenvectors with the $m$ largest contributions to the entropy, $\lambda_n \langle \mathbf{1},\mathbf{w}_n\rangle^2$. This is the method called Kernel Entropy Component Analysis [Jenssen2010] which can be seen to be an information-theoretic generalization of the PCA.

## 2.9 Robust PCA

There are many situations in which the input vectors **x** are contaminated by outliers. If this is the case, the outliers may dominate the estimation of the mean and covariance of the input data resulting in very poor performance of the PCA (see Fig. 11). There are approaches to have a robust PCA which basically amounts to have robust estimates of the mean and covariance.

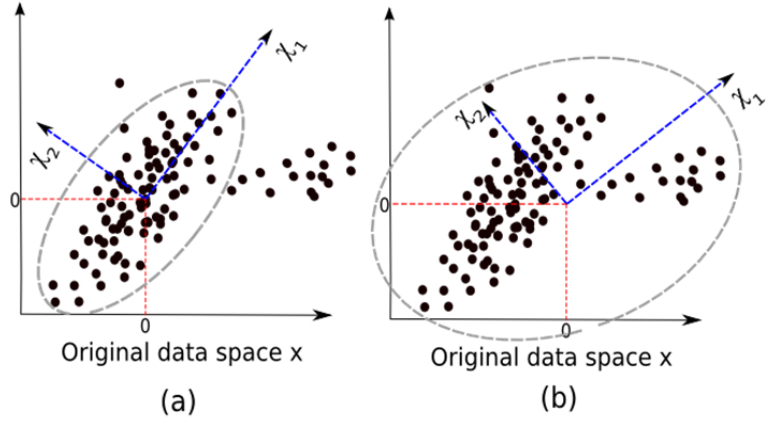

*Figure 11. Comparison between Robust PCA (a) and Standard PCA (b).*

An obvious modification to deal with univariate outliers is to change $l_2$-norm of the PCA objective function, $J_{PCA} = E\{\|\mathbf{x} - W\boldsymbol{\chi}\|^2\}$, by a $l_1$-norm which is known to be more robust to outliers, $J_{RPCA} = E\{\|\mathbf{x} - W\boldsymbol{\chi}\|_1\}$ [Baccini1996, Ke2005]. However, these modifications are not invariant to rotations of the input features [Ding2006]. [Ding2006] solved this problem by using the $R_1$ norm of a matrix that is defined as $\|E\|_{R_1} = \sum_{n=1}^{N}\left(\sum_{i=1}^{M} e_{in}^2\right)^{\frac{1}{2}}$, and constructing the objective function $J_{RPCA} = \|X - WW^t X\|_{R_1}$. Another possibility is to substitute the norm by a kernel as is done in robust statistics. This can be done at the level of individual variables $J_{RPCA} = E\left\{\sum_{i=1}^{M} k(\mathbf{x}_i - (W\boldsymbol{\chi})_i)\right\}$ (for instance [He2011] used a kernel based on the correntropy function) or at the level of the multivariate distance $J_{RPCA} = E\{k(\|\mathbf{x} - W\boldsymbol{\chi}\|^2)\}$ ([Iglesias2007] proposed to use the function $k(x) = x^\alpha$ with $\alpha$ about 0.5, they refer to this method as $\alpha$-PCA; [Subbarao2006] proposed to use M-estimators, they refer to their method as pbM (projection based M-estimators)).

[Kwak2008] proposes a different approach, instead of minimizing the L$_1$ norm of the reconstruction error as before $J_{RPCA} = E\{\|\mathbf{x} - W\boldsymbol{\chi}\|_1\}$, he maximizes the L$_1$ norm of the projections $J_{RPCA} = \|W^t X\|_1$ subject to $W^t W = I$ trying to maximize the dispersion of the new features.

The approach of De la Torre [Delatorre2003] can deal with univariate outliers. It modifies the PCA objective function to explicitly account for individual components of the observations that can be regarded as outliers and to account for the possible differences in the variance of each variable. The Robust PCA goal function is then $J_{RPCA} = \sum_{n=1}^{N}\sum_{i=1}^{M} O_{ni}\rho_i(\mathbf{x}_{ni} - (\boldsymbol{\mu}_i + \mathbf{w}_i\boldsymbol{\chi}_n)) + P(O_{ni})$, where $O_{ni}$ is a variable between 0 and 1 stating whether the $i$-th component of the $n$-th observation is an outlier (low values) or not (high values). If it is an outlier, then its error will not be counted, but the algorithm will be penalized by $P(O_{ni}) = (\sqrt{O_{ni}} - 1)^2$ to avoid the trivial solution of considering all samples to be an outlier. This algorithm does not assume that the input data is zero valued and estimates the mean, **μ**, from the non-outlier samples and components. For each component, the function $\rho_i$ robustly measures the error committed. This is done by $\rho_i(e) = \frac{e^2}{e^2 + \sigma_i^2}$, i.e., the squared error is "modulated" by the variance of that variable $\sigma_i^2$.

The approach of Hubert [Hubert2004] addresses the problem of multivariate outliers. It distinguishes the case when we have more observations than variables ($N > M$) and when we do not ($N < M$). In the first case ($N > M$), we have to specify the number $h$ of outliers we want the algorithm to be resistant to (it has to be $h < \frac{1}{2}(N - M - 1)$). Then, we look for the subset of $N - h$ input vectors such that the determinant of its covariance matrix is minimum (this determinant is a generalization of the variance to multivariate variables: when the determinant is large it means that the corresponding dataset has a large "variance"). We compute the average and covariance of this subset and make some adjustments to account for the finite-sample effect. These estimates are called the MCD (Minimum Covariance Determinant) estimates. Finally, PCA is performed as usual on this covariance estimate. In the second case ($N < M$), we cannot proceed as before since the determinant of the covariance matrix of any subset will be zero (remind our discussion when talking about the SVD decomposition of the covariance matrix). So we first perform a dimensionality reduction without loss of information using SVD and keeping $N - 1$ variables. Next, we identify outliers by choosing a large number of random directions, and projecting the input data onto each direction. For each direction, we compute MCD estimates (robust to $h$ outliers) of the mean ($\hat{\mu}_{MCD,\mathbf{w}}$) and standard deviation ($s_{MCD,\mathbf{w}}$) of the projections (note that these are univariate estimates). The outlyingness of each input observation is computed as $outl(\mathbf{x}_n) = \max_{\mathbf{w}} \frac{|\langle \mathbf{x}_n, \mathbf{w} \rangle - \hat{\mu}_{MCD,\mathbf{w}}|}{s_{MCD,\mathbf{w}}}$. The $h$ points with the highest outlyingness are removed from the dataset and, finally, PCA is performed normally on the remaining points.

[Pinto2011] proposes a totally different approach. It is well known that rank-statistics is more robust to noise and outliers than the standard statistical analysis. For this reason, they propose to substitute the original observations by their ranks (the i-th component of the n-th individual, $\mathbf{x}_{ni}$, is ranked among the i-th component of all individuals, then the observation $\mathbf{x}_{ni}$ is substituted by its rank that we will refer to as $\mathbf{r}_{ni}$, and the corresponding data matrix will be referred to as $R$). Looking at the PCA objective function, $J_{PCA} = \text{Tr}\{W^t \Sigma_X W\}$, they recognize that the covariance matrix in that equation, $\Sigma_X$, is very much related to the correlation coefficient among the different input features, the ij-th entry of this matrix is the covariance between the i-th and j-th variable. Therefore, they propose to substitute this covariance matrix by a new correlation measure better suited to handle rank data. They refer to the approach as Weighted PCA. The objective function is finally $J_{PCA} = \text{Tr}\{W^t \Sigma_R W\}$ subject to $W^t W = I$.

## 2.10 Factor Analysis

Factor analysis (FA) [Spearman1904, Thurstone1947, Kaiser1960, Lawley1971, Mulaik1971, Harman1976] is another statistical technique intimately related to PCA and dimensionality reduction. FA is a generative model that assumes that the observed data has been produced from a set of latent, unobserved variables (called factors) through the equation $\mathbf{x} = W\boldsymbol{\chi} + \mathbf{n}$ (if there is no noise, this model is the same as in PCA, although PCA is not a generative model in its conception). In this technique all the variances of the factors are absorbed into $W$ such that the covariance of $\boldsymbol{\chi}$ is the identity matrix. Factors are assumed to follow a multivariate normal distribution, and to be uncorrelated to noise. Under these conditions, the covariance of the observed variables is $C_\mathbf{x} = WW^t + C_\mathbf{n}$ where $C_\mathbf{n}$ is the covariance matrix of the noise and has to be estimated from the data. Matrix $W$ is solved by factorization of the matrix $WW^t = C_\mathbf{x} - C_\mathbf{n}$. This factorization is not unique since any orthogonal rotation of $W$ results in the same decomposition of $C_\mathbf{x} - C_\mathbf{n}$ [Kaiser1958]. This fact, rather than a drawback, is exploited to produce simpler factors in the same way as the PCA was rotated (actually the rotation methods for FA are the same as the ones for PCA).

## 2.11 Independent Component Analysis

Independent Component Analysis (ICA) [Common1994,Hyvarinen1999,Hyvarinen2000,Hyvarinen2001] is an example of information-theory based algorithm. It is also a generative model with equation $\mathbf{x} = W\chi$. While PCA looks for uncorrelated factors (i.e., a constraint on the second-order statistics), ICA looks for independent factors (i.e., a constraint on all their moments, not only second-order). This is an advantage if the factors are truly independent (for two Gaussian variables decorrelation implies independence but this is not generally true for variables with any other distribution). In the language of ICA, the $\chi$ vectors are called the sources, while the $\mathbf{x}$ are called the observations. Matrix $W$ is called the mixing matrix and the problem is formulated as one of source separation, i.e., finding an unmixing matrix $\tilde{W}$ such that the components of $\hat{\chi} = \tilde{W}^t \mathbf{x}$ are as independent as possible (see Fig. 12). The sources can be recovered up to a permutation (the sources are recovered in different order) and a scale change (the sources are recovered with different scale). A limitation of the technique is that usually sources are supposed to be non-Gaussian since the linear combination of two Gaussian variables is also Gaussian making the separation an ill-posed problem.

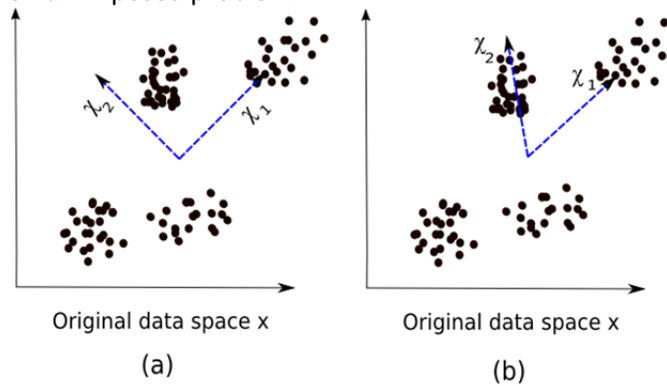

*Figure 12. a) Example of PCA results for a given input distribution. b) ICA results for the same distribution.*

Different ICA methods differ in the way they measure the independence of the estimates of the source variables, resulting in different estimates of the mixing matrix $W$ and source variables $\hat{\chi}$. The following are different options commonly adopted:

- Non-Gaussianity: The central limit theorem states that the distribution of the weighted sum of the sources tends to be normally distributed whichever the distributions of the original sources. Thus, a possible way to achieve the separation of the sources is by looking for transformations $\tilde{W}$ that maximize the kurtosis of each of the components of the vector $\hat{\chi}$ [Hyvarinen2001] (see Fig. 13). The kurtosis is related to the third order moment of the distribution. The kurtosis of the Gaussian is zero and, thus, maximizing the kurtosis, minimizes the Gaussianity of the output variables. In fact, maximizing the kurtosis can be seen as a Non-linear PCA problem (see above) with the non-linear function for each feature vector $f_i(\chi_i) = \chi_i + \text{sgn}(\chi_i)\chi_i^2$. FastICA is an algorithm based on this goal function. The problem of kurtosis is that it can be very sensitive to outliers. Alternatively, we can measure the non-Gaussianity by negentropy which is the Kullack-Leibler divergence between the multivariate distribution of the estimated sources $\hat{\chi}$, and the distribution of a multivariate variable $\chi_G$ of the same mean and covariance matrix as $\hat{\chi}$. Projection pursuit [Friedman1974, Friedman1987] is an exploratory data analysis algorithm looking for projection directions of maximum kurtosis (as in our first ICA algorithm) while Exploratory Projection Pursuit [Girolami1997] uses the maximum negentropy to look for the projection directions. Non-Gaussian Component Analysis [Theis2011], instead of looking for a single direction like in projection pursuit, looks for an entire linear subspace where the projected data is as non-Gaussian as possible.

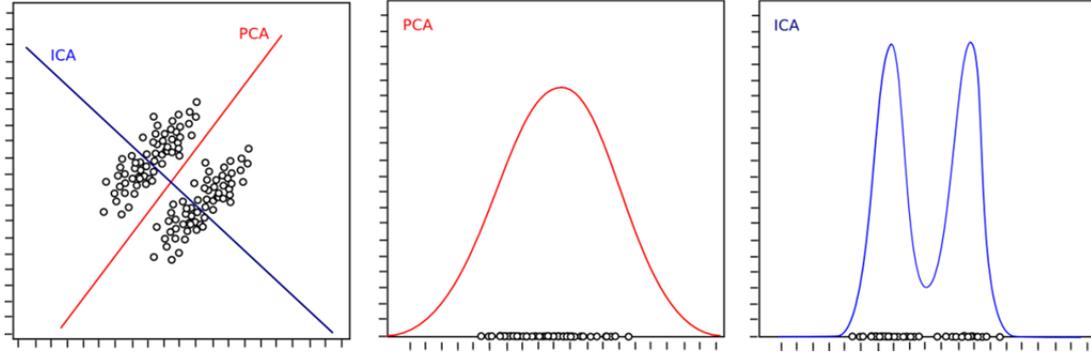

*Figure 13. a) Sample distribution in which the first PCA component maximizes the explained variance but projections onto this direction are normally distributed (b). ICA first component is also shown on the sample data, data projection onto this direction clearly shows a non-Gaussian distribution.*

- Maximum-likelihood (ML): Let us assume that we know the *a priori* distribution of each of the components $\hat{\chi}$. Then, we could find the matrix $\tilde{W}$ simply by maximizing the likelihood of all the observations $l(\mathbf{x}) = p_\mathbf{x}(\mathbf{x}) = |\det \tilde{W}| p_\chi(\hat{\chi}) = |\det \tilde{W}| \prod_{i=1}^{m} p_{\chi_i}(\hat{\chi}_i)$. Taking logarithms and the expected value over all input vectors we obtain $L(X) = \log|\det \tilde{W}| + E\left\{\sum_{i=1}^{m} p_{\chi_i}(\hat{\chi}_i))\right\}$ which is the objective function to maximize with respect to $\tilde{W}$ (the Bell-Sejnowski and the natural gradient algorithms [Hyvarinen2001] are typical algorithms for performing this optimization). If the distribution of the features is not known *a priori*, it can be shown [Hyvarinen2001] that "reasonable" errors in the estimation of the $p_{\chi_i}$ distributions result into locally consistent ML estimators as long as for all $i$ $E\{\chi_i g_i(\chi_i) - g_i'(\chi_i)\} > 0$, where $g_i(\chi_i) = \frac{\tilde{p}_{\chi_i}'(\chi_i)}{\tilde{p}_{\chi_i}(\chi_i)}$ and $\tilde{p}_{\chi_i}$ is our estimate of the truly underlying distribution $p_{\chi_i}$. A common approach is to choose for each $i$ between a super-Gaussian and a sub-Gaussian distribution. This choice is done by checking which one of the two fulfills $E\{\chi_i g_i(\chi_i) - g_i'(\chi_i)\} > 0$. Common choices are $g_i(\chi_i) = \chi_i + \tanh(\chi_i)$ for the super-Gaussian and $g_i(\chi_i) = \chi_i - \tanh(\chi_i)$ for the sub-Gaussian. This criterion is strongly related to the Infomax (Information Maximization) in neural networks. There, the problem is to look for the network weights such that the joint entropy of the output variables, $H(\chi_1, \chi_2, ..., \chi_m)$, is maximized. It has been shown [Lee2000] that the Infomax criterion is equivalent to the maximum likelihood one when $g_i(\chi_i)$ is the non-linear function of each output neuron of the neural network (usually a sigmoid). Equivalently, instead of maximizing the joint entropy of the feature variables, we could have minimized their mutual information. Mutual information is a measure of the dependency among a set of variables. So, it can be easily seen how all these criteria is maximizing the independence of the feature vectors.

- Nonlinear decorrelation: two variables $\chi_1$ and $\chi_2$ are independent if for all continuous functions $f_1$ and $f_2$ with finite support we have $E\{f_1(\chi_1) f_2(\chi_2)\} = E\{f_1(\chi_1)\} E\{f_2(\chi_2)\}$. The two variables are non-linearly decorrelated if $E\{f_1(\chi_1) f_2(\chi_2)\} = 0$. In fact, PCA looks for output variables that are second-order decorrelated, $E\{\chi_1 \chi_2\} = 0$, although they may not be independent because of their higher-order moments. Making the Taylor expansion of $f_1(\chi_1)f_2(\chi_2)$ we arrive to $E\{f_1(\chi_1)f_2(\chi_2)\} = \sum_{i,j=0}^{\infty} c_{ij} E\{\chi_1^i \chi_2^j\} = 0$. For which we need that all high-order correlations are zero, $E\{\chi_1^i \chi_2^j\} = 0$. Hérault-Jutten and Cichocki-Unbehauen algorithms look for independent components by making specific choices of the non-linear functions $f_1$ and $f_2$ [Hyvarinen2001].

## 3. Methods based on Dictionaries

Another family of methods is based on the decomposition of a matrix formed by all input data as columns, $X$. The input data matrix using the input variables is transformed into a new data matrix using the new variables. The transformation is nothing but a linear change of basis between the two variable sets. In the field of dimensionality reduction, the matrix expressing the change of basis is known as a dictionary (dictionary elements are called atoms) and there are several ways of producing such a dictionary [Rubinstein2010]. We have already seen Singular Value Decomposition (SVD) and its strong relationship to PCA. Vector Quantization (K-means) can also be considered an extreme case of dictionary based algorithm (input vectors are represented by a single atom, instead of as a combination of several atoms). Under this section we will explore other dictionary based algorithms.

### 3.1 Non-negative Matrix Factorization

A drawback of many dimensionality reduction techniques is that they produce feature vectors with negative components. In some applications like text analysis, it is natural to think of the observed vectors as the weighted sum of some underlying "factors" with no subtraction involved (for instance, it is natural to think that if a scientific article is about two related topics, the word frequencies of the article will be a weighted sum (with positive weights) of the word frequencies of the two topics). Let us consider the standard decomposition dictionary $X = WU$ where $W$ is the dictionary (of size $M \times m$, its columns are called "atoms" of the dictionary) and $U$ (of size $m \times N$) is the expression of the observations in the subspace spanned by the dictionary atoms (see Fig. 14). Not considering subtractions imply constraining the feature vectors to be $U > 0$.

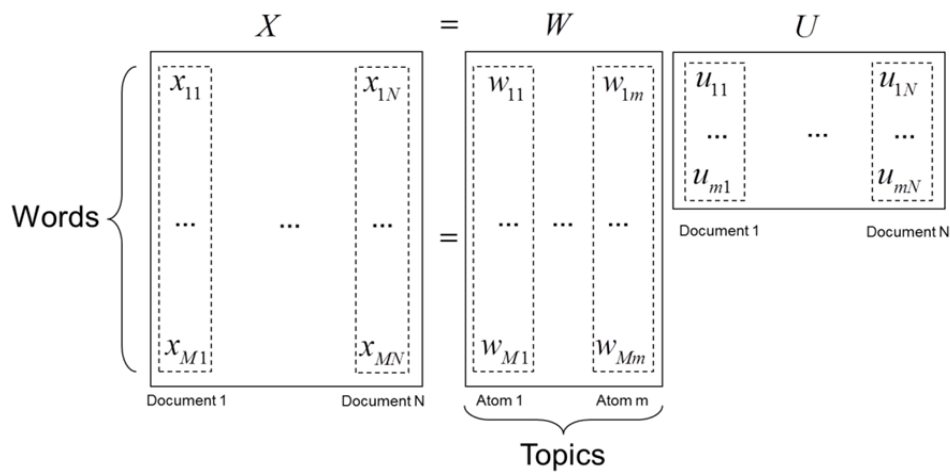

*Figure 14. Dictionary decomposition of a set of documents (see Fig. 3). Each document is decomposed as the linear combination given by the weights in U of the topics (atoms) contained in W.*

If $X$ is made only of positive values, it might be interesting to constrain the atoms to be positive as well ($W > 0$). This is the problem solved by Non-negative Matrix Factorization [Lee1999, Lee2001]. The goal is to minimize $\|X - WU\|_F^2$ or $D(X\|WU)$ (defined as $D(A\|B) = \sum_{i,j}\left(A_{ij}\log\frac{A_{ij}}{B_{ij}} - A_{ij} + B_{ij}\right)$; this is the Kullback-Leibler divergence if $A$ and $B$ are normalized so that they can be regarded as probability distributions) subject to $W, U > 0$. The advantage of this decomposition is that, if the application is naturally defined with positive values, the dictionary atoms are much more understandable and related to the problem than the standard dimensionality reduction methods. [Sandler2011] proposed to minimize the Earth Mover's Distance between the matrices $X$ and $WU$ with the aim of making the method more robust, especially to small samples. The Earth's Mover Distance, also called Wasserstein metric, is a way of measuring distances between two probability distributions (for a review on how to measure distances between probability distributions see [Rubner2000]). It is defined as the minimum cost of turning one probability distribution into the other and it is computed through a transportation problem. This distance was extended by [Sandler2011] to measure the distance between matrices by applying the distance to each column (feature) of the matrices and then summing all distances.

In the recent years there is much interest in the construction of sparse feature vectors, sparse dictionaries or both. The underlying idea is to produce feature vectors with as many zeroes as possible, or what is the same, approximating the observations with as few dictionary atoms as possible. This has obvious advantages when trying to explain the "atomic composition" of a given observation. In the following paragraphs we will review some of the approaches already proposed for NMF in this direction.

Local NMF [Feng2002] enforces sparsity by adding to the NMF goal function the term $\|W^t W\|_1$ (which promotes the orthogonality of the atoms, i.e., minimizes the overlapping between atoms) and $-\|U\|_2^2$ (which maximizes the variance of the feature vectors, i.e., it favours the existence of large components). Non-negative sparse coding [Hoyer2002] and Sparse NMF [Liu2003] add the term $\|U\|_1$ in order to minimize the number of features with significant values. Pauca *et al.* [Pauca2006] regularize by adding $\|U\|_2^2$ and $\|W\|_2^2$ (this is especially suited for noisy data). NMF with Sparseness Constraints [Hoyer2004] performs the NMF with the constraint that the sparseness of each column of $W$ is a given constant $S_W$ (i.e., it promotes the sparseness of the dictionary atoms) and the sparseness of each row of $U$ is another constant $S_U$ (i.e., it promotes that each atom is used in as few feature vectors as possible). In [Hoyer2004], the sparseness of a vector is defined as $Sparseness(\mathbf{x}) = \frac{1}{\sqrt{n}-1}\left(\sqrt{n} - \frac{\|\mathbf{x}\|_1}{\|\mathbf{x}\|_2}\right)$ that measures how much energy of the vector is packed in as few components as possible. This function evaluates to 1 if there is a single non-zero component, and evaluates to 0 if all the elements are identical. Non-smooth NMF [Pascual-Montano2006] modifies the NMF factorization to $X = WS_\lambda U$. The matrix $S_\lambda$ controls the sparseness of the solution through the parameter $\lambda$. It is defined as $S_\lambda = (1-\lambda)I + \lambda \frac{1}{m}\mathbf{1}\mathbf{1}^t$. For $\lambda = 0$ it is the standard NMF. For $\lambda = 1$, we can think of the algorithm as using "effective" feature vectors defined by $U_\lambda = S_\lambda U$ that substitute each feature vector $\chi$ by a vector of the same dimensionality whose all components are equal to the mean of $\chi$. This is just imposing non-sparseness on the feature vectors, and this will promote sparseness on the dictionary atoms. On the other hand, we could have also thought of the algorithm as using the "effective" atoms given by $W_\lambda = WS_\lambda$ that substitute each atom by the average of all atoms. In this case, the non-sparseness of the dictionary atoms will promote sparseness of the feature vectors. Non-smooth NMF is used with typical $\lambda$ values about 0.5.

Another flavor of NMF enforces learning the local manifold structure of the input data [Cai2011b], Graph-regularized NMF (GNMF). Assuming that the input vectors $\mathbf{x}_i$ and $\mathbf{x}_j$ are close in the original space, one might like that the reduced representations, $\chi_i$ and $\chi_j$, are also close. For doing so, the algorithm constructs a graph $G$ encoding the neighbors of the input observations. Observations are represented by nodes in the graph, and two nodes are connected by an edge if their distance is smaller than a given threshold and they are in the K-neighbors list of each other. The weight of each edge is 1, or if we prefer we can assign a different weight to each edge depending on the distance between the two points (for instance, $e^{-\frac{1}{2}\|\mathbf{x}_i - \mathbf{x}_j\|^2}$). We build the diagonal matrix $D$ whose elements are the row sums of $G$. The Laplacian of this graph is defined as $L = D - G$. The sum of the Euclidean distances of the reduced representations corresponding to all neighbor pairs can be computed as $\text{Tr}\{ULU^t\}$ [Cai2011b]. In this way, the GNMF objective function becomes $\|X - WU\|_F^2 + \lambda \text{Tr}\{ULU^t\}$. This algorithm has the corresponding version in case that the Kullback-Leibler divergence is preferred over Euclidean distances [Cai2011b].

## 3.2 Principal Tensor Analysis and Non-negative Tensor Factorization

In some situations the data is better represented by a multidimensional array rather than by a matrix. For instance, we might have a table representing the gene expression level for a number of drugs. It is naturally represented by a (drug, gene) matrix and all the previous methods to factorize matrices are applicable. However, we might have a (drug, gene, time) table that specifies the gene expression for each combination of gene, drug and time. This three-way table (and in general multiway tables) is a tensor (strictly speaking a multiway table of dimension $d$ is a tensor if and only if it can be decomposed as the outer product of $d$ vectors; however, in the literature multiway tables are usually referred to as tensors and we will also adhere here to this loose definition). Tensors can be flattened into matrices and then all the previous techniques would be available. However, the locality imposed by some variables (like time or spatial location) would be lost. Non-negative tensor factorization [Cichocki2009] is an extension of NMF to multiway tables. The objective is, as usual, minimizing the representation error $\left\|X - \sum_{i=1}^{m} \mathbf{w}_i^1 \otimes \mathbf{w}_i^2 \otimes ... \otimes \mathbf{w}_i^d\right\|_F^2 = \left\|X - \sum_{i=1}^{m}\bigotimes_{j=1}^{d} \mathbf{w}_i^j\right\|_F^2$ subject to $\mathbf{w}_i^j \geq 0$ for all $i$ and $j$. In the previous expression $X$ is a tensor of dimension $d$ (in our three-way table example, $d = 3$), $\otimes$ represents the outer product, $m$ is a parameter controlling the dimensionality reduction. For each dimension $j$ (drug, gene or time, in our example), there will be $m$ associated vectors $\mathbf{w}_i^j$. The length of each vector depends on the dimension it is associated to (see Fig. 15). If there are $N_j$ elements in the $j$-th dimension ($N_{drugs}$, $N_{genes}$ and $N_{time}$ entries in our example), the length of the vectors $\mathbf{w}_i^j$ is $N_j$. The approximation after dimensionality reduction is $\hat{X} = \sum_{i=1}^{m}\bigotimes_{j=1}^{d} \mathbf{w}_i^j$. For a three-way table, the $pqr$ element of the tensor is given by $\hat{X}_{pqr} = \sum_{i=1}^{m} \mathbf{w}_{ip}^1 \mathbf{w}_{iq}^2 \mathbf{w}_{ir}^3$, where $\mathbf{w}_{ip}^1$ represents the $p$-th component of the vector $\mathbf{w}_i^1$ (analogously with $\mathbf{w}_{iq}^2$ and $\mathbf{w}_{ir}^3$).

NTF is a particular case of a family of algorithms decomposing tensors. PARAFAC [Harshman1970, Bro1997] may be one of the first algorithms doing so and can be regarded as a generalization of SVD to tensors. The SVD approximation $\hat{X} = W_m D_m U_m$ can be rewritten as $\hat{X} = \sum_{i=1}^{m} d_i \mathbf{w}_i \mathbf{u}_i^t = \sum_{i=1}^{m} d_i \mathbf{w}_i \otimes \mathbf{u}_i$. PARAFAC model is minimizing the approximation error $J_{PARAFAC} = \left\|X - \sum_{i=1}^{m} d_i \mathbf{w}_i^1 \otimes \mathbf{w}_i^2 \otimes ... \otimes \mathbf{w}_i^d\right\|_F^2$. We can enrich the model to consider more outer products as in $J_{Tucker} = \left\|X - \sum_{i_1=1}^{m_1}\sum_{i_2=1}^{m_2}...\sum_{i_d=1}^{m_d} d_{i_1 i_2 ... i_d} \mathbf{w}_{i_1}^1 \otimes \mathbf{w}_{i_2}^2 \otimes ... \otimes \mathbf{w}_{i_d}^d\right\|_F^2$. This is called the Tucker model [Tucker1966].

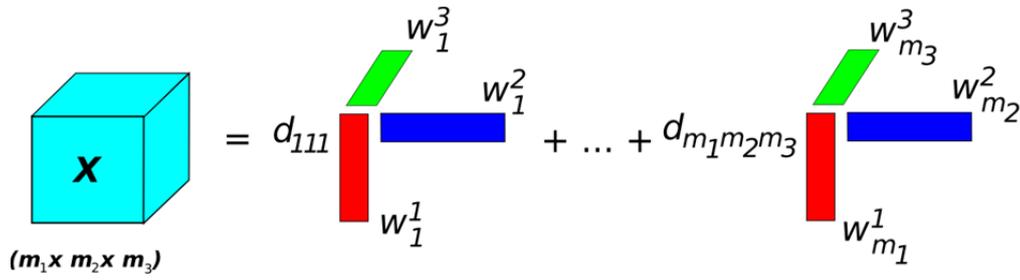

*Figure 15. Tensor decomposition as the sum of a series of outer vector products. The figure corresponds to the more general Tucker model, which is simplified to PARAFAC or NTF.*

## 3.3 Generalized SVD

According to the SVD decomposition, $X = WDU$, the columns of $W$ are the eigenvectors of the matrix $XX^t$, while the rows of $U$ are the eigenvectors of the matrix $X^tX$. Eigenvectors are orthogonal and therefore $W^tW = I$ and $UU^t = I$. After performing the dimensionality reduction, the approximation $\hat{X} = W_m D_m U_m$ is the matrix minimizing the representation error $\varepsilon = \|X - \hat{X}\|^2 = trace\{(X-\hat{X})(X-\hat{X})^t\}$.

Generalized SVD [Paige1981] performs a similar decomposition but relaxes, or modifies, the orthogonality conditions according to two constrain matrices, $C_W$ and $C_U$, such that $W^t C_W W = I$ and $U C_U U^t = I$ [Abdi2007]. After the dimensionality reduction, the approximation matrix is the one minimizing $\varepsilon = trace\{C_W(X-\hat{X})C_U(X-\hat{X})^t\}$.

Generalized SVD is a very versatile tool since under the appropriate choices of the constrain matrices it can be particularized to correspondence analysis (a generalization of factor analysis for categorical variables, $C_W$ is the relative frequency of the rows of the data matrix, $X$, and $C_U$ is the relative frequencies of its rows), discriminant analysis (a technique relating a set of continuous variables to a categorical variable), and canonical correlation analysis (a technique analyzing two groups of continuous variables and performing simultaneously two dimensionality reductions so that the two new sets of features have maximum cross correlation) [Abdi2007].

## 3.4 Sparse representations and overcomplete dictionaries

A different approach to dimensionality reduction is by cutting the input signal into small pieces, and performing a dimensionality reduction of them. For instance, we can divide an image into small 8x8 pieces (vectors of dimension 64). Then we try to express each piece as a linear combination of a few atoms from a large dictionary (of size larger than 64, that is why it is called overcomplete). At the level of pieces, the dictionary acts as a dimensionality expansion, although overall, there is a dimensionality reduction since each piece can be represented with just a few atoms instead of the 64 values needed originally. This approach can be applied to any domain where the original vectors $\mathbf{x}$ can be decomposed into pieces of similar nature: images and time series are good examples. Let us call $\hat{\mathbf{x}}$ to each one of these pieces. The idea is that for each piece we solve the problem $\min \|\hat{\boldsymbol{\chi}}\|_0$ subject to $\|\hat{\mathbf{x}} - \hat{W}\hat{\boldsymbol{\chi}}\|_2 \leq \varepsilon$ (by setting $\varepsilon = 0$ we require exact representation of the original vector). The columns of $\hat{W}$ are the atoms, and the feature vector $\hat{\boldsymbol{\chi}}$ define the specific linear combination of atoms used to represent the corresponding piece. The sparseness of the feature vector is measured simply by counting the number of elements different from zero (this is usually referred to as the $l_0$ norm, although actually it is not a norm since it is not positive homogeneous). The problem of the $l_0$ norm is that it yields non-convex optimization problems whose solution is NP-complete. For tackling this problem some authors have replaced the $l_0$ by the $l_p$ norm (for $0 < p < 1$ the problem is still non-convex, although there are efficient algorithms; $p = 1$ is a very popular norm to promote sparseness). Related problems are $\min \|\hat{\mathbf{x}} - \widehat{W}\widehat{\boldsymbol{\chi}}\|_2 + \lambda \|\widehat{\boldsymbol{\chi}}\|_p$ (the Least Absolute Shrinkage and Selector Operator (LASSO) is such a problem with $p = 1$, and ridge regression is also this problem with $p = 2$) and $\min \|\hat{\mathbf{x}} - \widehat{W}\widehat{\boldsymbol{\chi}}\|_2$ subject to $\|\widehat{\boldsymbol{\chi}}\|_p \leq t$ where $t$ is a user-defined value (for $p = 0$ it restricts the feature vector to use at most $t$ atoms). The previous problems are called the sparse coding step and many algorithms have been devised for its solution. The most popular ones are basis pursuit [Chen1994, Chen2001], matching pursuit [Mallat1993], orthogonal matching pursuit [Pati1993, Tropp2007], orthogonal least squares [Chen1991], focal underdetermined system solver (FOCUSS) [Gorodnitsky1997], gradient pursuit [Blumensath2008] and conjugate gradient pursuit [Blumensath2008]. For a review on these techniques, please, see [Bruckstein2009].

The other problem is how to learn the overcomplete dictionary $\widehat{W}$ from the input pieces $\hat{\mathbf{x}}$. In a way, this can be considered as an extension of the vector quantization problem. In vector quantization we look for a set of class averages minimizing the representation error when all except one of the feature vector components are zero (the value of the non-null $\chi_i$ component is 1). Now, we have relaxed this condition and we allow the feature components to be real valued (instead of 0 or 1) and we represent our data by a weighted sum of a few atoms. Nearly all methods iteratively alternate between the estimation of the feature vectors and the estimation of the dictionary, and they differ in the goal function being optimized, which ultimately result into different update equations for the dictionary. The Method of Optimal Directions (MOD) [Engan2000] is a possible way of learning dictionaries. This method optimizes $\widehat{W}^*, \widehat{U}^* = \arg\min_{W,U} \|X - \widehat{W}\widehat{U}\|_F^2$ and it has proven to be very efficient. Another possibility to learn the dictionary is by Maximum Likelihood [Lewicki2000]. Under a generative model it is assumed that the observations have been generated as noisy versions of a linear combination of atoms $\hat{\mathbf{x}} = \widehat{W}\widehat{\boldsymbol{\chi}} + \boldsymbol{\varepsilon}$. The observations are assumed to be produced independently, the noise to be white and Gaussian, and the *a priori* distribution of the feature vectors to be Cauchy or Laplacian. Under these assumptions the problem of maximizing the likelihood of observing all pieces, $\hat{\mathbf{x}}_n$, given $\widehat{W}$ is maximized by $\widehat{W}^* = \arg\min_{\widehat{W}} \sum_n \min_{\chi_n} \left\{ \|\hat{\mathbf{x}}_n - \widehat{W}\widehat{\boldsymbol{\chi}}_n\|_2^2 + \lambda \|\widehat{\boldsymbol{\chi}}_n\|_1 \right\}$. If a prior distribution of the dictionary is available we can use a Bayesian approach (Maximum *a posteriori*) [Kreutz2003]. K-SVD [Aharon2006] solves the problem $\widehat{W}^*, \widehat{U}^* = \arg\min_{\widehat{W},\widehat{U}} \|X - \widehat{W}\widehat{U}\|_F^2$ subject to $\|\widehat{\boldsymbol{\chi}}_n\|_1 \leq t$ for all the pieces, $n$. It is conceived as generalization of the K-means algorithm, and in the update of the dictionary there is a Singular Value Decomposition (therefore, its name).

K-SVD can be integrated into a larger framework capable of optimally reconstructing the original vector from its pieces. The goal function is $\lambda \|\mathbf{x} - \hat{\mathbf{x}}\|_2^2 + \sum_n \left( \lambda_n \|\widehat{\boldsymbol{\chi}}_n\|_0 + \|\hat{\mathbf{x}}_n - \widehat{W}\widehat{\boldsymbol{\chi}}_n\|_2^2 \right)$, where the $\lambda_n$ variables are Lagrangian multipliers. Once all the patches have been approximated by their corresponding feature vectors, we can recover the original input vector by $\hat{\mathbf{x}} = \left( \lambda I + \sum_n P_n^t P_n \right)^{-1} \left( \lambda \mathbf{x} + \sum_n P_n^t \widehat{W}\widehat{\boldsymbol{\chi}}_n \right)$, where $P_n$ is an operator extracting the $n$-th piece as a vector, and $P_n^t$ is the operator putting it back in its original position.

Tensor representations can be coupled to sparse representations as shown by [Gurumoorthy2010] which we will refer to as Sparse Tensor SVD. In certain situations, the input data is better represented by a matrix or tensor than by a vector. For instance, image patches can be represented as a vector by lexicographically ordering the pixel values. However, this representation spoils the spatial correlation of nearby pixels. Let us then consider that we no longer have input vectors, $\mathbf{x}_n$, but input matrices (we will generalize later to tensors). This method learns a dictionary of $K$ SVD-like basis $(W_i, U_i)$. Each input matrix, $X_n$, is sparsely represented in the $i$-th SVD-like basis, $X_n \approx W_i D_n U_i$. Sparsity is measured through the number of non-zero components of the representation $\|D_n\|_0$. Finally, a membership matrix, $\widehat{u}_{in} \in (0,1)$, specifies between 0 (no membership) and 1 (full membership) whether the input matrix $X_n$ is represented with the $i$-th basis or not. In this way, the goal of this algorithm is the minimization of the representation error given by $J_{SparseTensorSVD} = \sum_{n=1}^{N} \sum_{i=1}^{K} \widehat{u}_{in} \|X_n - W_i D_n U_i\|_F^2$. This functional is minimized with respect to the SVD-like basis, the membership function and the sparse representations. The optimization is constrained by the orthogonality of the basis ($W_i^t W_i = I$, $U_i U_i^t = I$, for all $i$), the sparsity constraints ($\|D_n\|_0 \leq t$, $t$ is a user-defined threshold), and that the columns of the membership matrix define a probability distribution ($\sum_{i=1}^{K} \widehat{u}_{in} = 1$). The generalization of this approach to tensors is straightforward simply replacing the objective function by $J_{SparseTensorSVD} = \sum_{n=1}^{N} \sum_{i=1}^{K} \widehat{u}_{in} \left\| X_n - U_n \otimes \left( \bigotimes_{j=1}^{d} \mathbf{w}_i^j \right) \right\|_F^2$ subject to the same orthogonality, sparsity and membership constraints.

## 4. Methods based on projections

A different family of algorithms poses the dimensionality reduction problem as one of projecting the original data onto a subspace with some interesting properties.

### 4.1 Projection onto interesting directions

Projection pursuit defines the output subspace by looking for "interesting" directions. What is "interesting" depends on the specific problem but usually directions in which the projected values are non-Gaussian are considered to be interesting. Projection pursuit looks for directions maximizing the kurtosis of the projected values as a measure of non-Gaussianity. This algorithm was visited during our review of Independent Component Analysis and presented as a special case of that family of techniques.

All the techniques presented so far are relatively costly in computational terms. Depending on the application it might be enough to reduce the dimensionality without optimizing any goal function but in a very fast way. Most techniques project the observations $\mathbf{x}$ onto the subspace spanned by a set of orthogonal vectors. However, choosing the best (in some sense) orthogonal vectors is what is computationally costly while the projection itself is rather quick. In certain application domains some "preconceived" directions are known. This is the case of the Discrete Cosine Transform (DCT) used in the image standard JPEG [Watson1993]. The "cosine" vectors usually yield good reduction results with low representation error for signals and images. Many other transform-based compression methods, like wavelets, also fall under this category. Alternatively, random mapping [Kaski1998, Dasgupta2000, Bingham2001] solves this problem by choosing zero-mean random vectors as the "interesting" directions onto which project the original observations (this amounts to simply taking a random matrix $W$ whose columns are normalized to have unit module). Random vectors are orthogonal in theory ($E\{\langle \mathbf{w}_i, \mathbf{w}_j \rangle\} = 0$), and nearly orthogonal in practice. Therefore, the dot product between any pair of observations is nearly conserved in the feature space. This is a rather interesting property since in many applications the similarity between two observations is computed through the dot product of the corresponding vectors. In this way, these similarities are nearly conserved in the feature space but at a computational cost that is only a small fraction of the cost of most dimensionality reduction techniques.

## 4.2 Projection onto manifolds

Instead of projecting the input data onto an "intelligent" set of directions, we might look for a manifold close to the data, project onto that manifold and unfold the manifold for representation. This family of algorithms is represented by a number of methods like Sammon projection, Multidimensional Scaling, Isomap, Laplacian eigenmaps, and Local linear embedding. All these algorithms can be shown to be special cases of Kernel PCA under certain circumstances [Williams2002, Bengio2004]. The reader is also referred to [Cayton2005] for an excellent review of manifold learning.

We already saw MDS as a technique preserving the inner product of the input vectors. Alternatively, it could be also seen as a technique preserving distances in the original space. From this point of view, Sammon projection [Sammon1969] and MDS [Kruskal1986] look for a projected set of points, $\chi$, such that distances in the output subspace are as close as possible to the distances in the original space. Let $D_{n_1 n_2} = d(\mathbf{x}_{n_1}, \mathbf{x}_{n_2})$ be the distances between any pair of observations in the original space (for an extensive review of distances see [Ramanan2011]). Let $d_{n_1 n_2} = d(\chi_{n_1}, \chi_{n_2})$ be the distance between their feature vectors, and $\hat{D}_{n_1 n_2} = d(\hat{\mathbf{x}}_{n_1}, \hat{\mathbf{x}}_{n_2})$ the distance between their approximations after dimensionality reduction. Classical MDS look for the feature vectors minimizing $\sum_{n_1, n_2} \left( D_{n_1, n_2} - \hat{D}_{n_1, n_2} \right)^2$, while Sammon projection minimizes $\dfrac{\sum_{n_1, n_2} \omega_{n_1, n_2} \left( D_{n_1, n_2} - d_{n_1, n_2} \right)^2}{\sum_{n_1, n_2} D_{n_1, n_2}^2}$ with $\omega_{n_1, n_2} = \dfrac{1}{D_{n_1, n_2}}$ (this goal function is called the stress function and it is defined between 0 and 1) (see Fig. 16).

Classical MDS solves the problem in a single step involving the computation of the eigenvalues of a Gram matrix involving the distances in the original space. In fact, it can be proved that classical MDS is equivalent to PCA [Williams2002]. Sammon projection uses a gradient descent algorithm and can be shown [Williams2002] to be equivalent to Kernel PCA. Metric MDS modifies the $D_{n_1 n_2}$ distances by an increasing, monotonic nonlinear function $f(D_{n_1 n_2})$. This modification can be part of the algorithm itself (some proposals are [Bengio2004] $f(D_{n_1 n_2}) = D_{n_1 n_2}^{\alpha}$ and $f(D_{n_1 n_2}) = \tfrac{1}{2}\left(D_{n_1 n_2} - D_{n_1 \cdot} - D_{\cdot n_2} + D_{\cdot \cdot}\right)$ where $D_{n_1 \cdot}$ and $D_{\cdot n_2}$ denote the average distance of observations $n_1$ and $n_2$ to the rest of observations, and $D_{\cdot \cdot}$ is the average distance between all observations) or as part of the data collection process (e.g. distances have not been directly observed but are asked to a number of human observers). If the distances $D_{n_1 n_2}$ are measured as geodesic distances (the geodesic distance between two points in a manifold is the one measured along the manifold itself; in practical terms it is computed as the shortest path in a neighborhood graph connecting each observation to its K-nearest neighbors, see Fig. 17), then the MDS method is called Isomap [Tenenbaum2000].

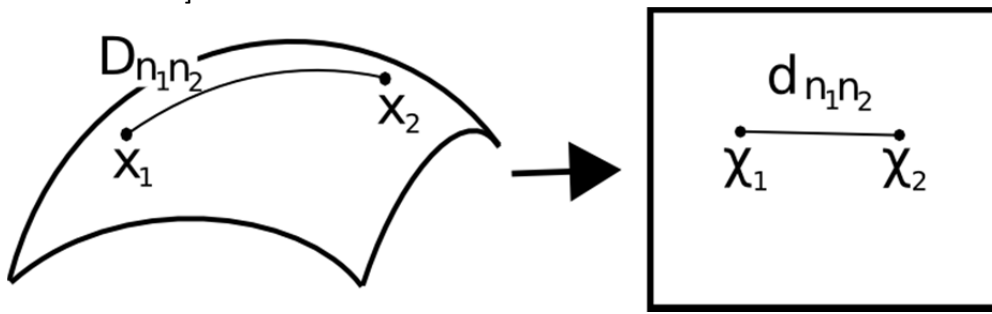

Fig.16. Distances in the original space are mapped onto the lowest dimensional space trying to find projection points that keep the set of distances as faithful as possible.

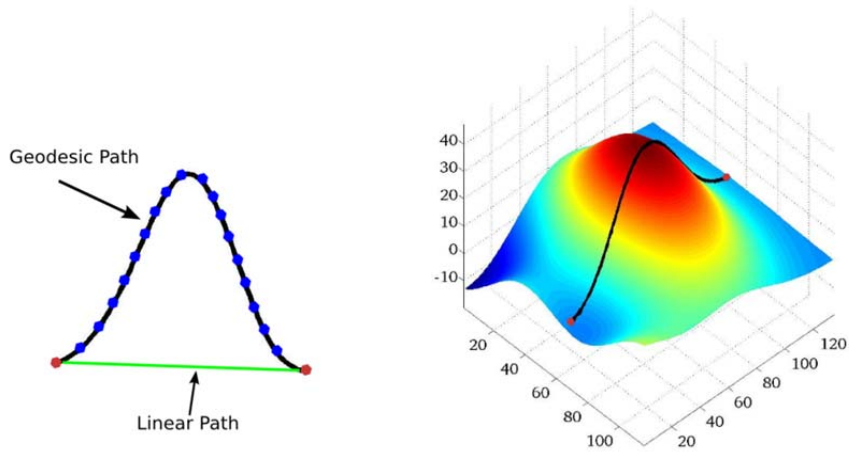

Figure 17. Geodesic versus Euclidean distance. The geodesic distance between two points is the length of the path belonging to a given manifold that joins the two points, while the Euclidean distance is the length of the linear path joining the two points.

Laplacian eigenmaps [Belkin2001, Belkin2002] start from an adjacency graph similar to that of Isomap for the computation of the geodesic distances. The neighbor similarity graph $G$ is calculated as was done in GNMF. Then, the generalized eigenvalues and eigenvectors of the Laplacian of the graph $G$ are computed, i.e., we solve the problem $(D-G)\mathbf{w} = \lambda D\mathbf{w}$. Finally, we keep the eigenvectors of the $m$ smallest eigenvalues discarding the smallest one which is always 0. The dimensionality reduction is performed by $\chi_n = (\mathbf{w}_{1n}, \mathbf{w}_{2n}, ..., \mathbf{w}_{mn})$, i.e., by keeping the $n$-th component of the $m$ eigenvectors. The interesting property of the Laplacian eigenmap is that the cost function, which measures the distance among the projected features, can be expressed in terms of the Graph Laplacian: $J_{LE} = \frac{1}{2}\sum_{i,j} G_{ij} \|\chi_i - \chi_j\|^2 = \chi_i^t L \chi_j$. So the goal is to minimize $J_{LE}$ subject to $\chi_i^t D \chi_i = 1$ [Zhang2009]. Finally, it is worth mentioning that Laplacian Eigenmaps and Principal Component Analysis (and their kernel versions) have been found to be particular cases of a more general problem called Least Squares Weighted Kernel Reduced Ranked Regression (LS-WKRRR) [Delatorre2012] (in fact, this framework also generalizes Canonical Correlation Analysis, Linear Discriminant Analysis and Spectral Clustering, techniques that are out of the scope of this review). The objective function is to minimize $J_{LS-WKRRR} = \|W_\chi (\Phi_\chi - BA^t \Phi_\mathbf{x}) W_\mathbf{x}\|_F^2$ subject to $\text{rank}(BA^t) = m$. $W_\chi$ is a diagonal weight matrix for the feature points, $W_\mathbf{x}$ is a diagonal weight matrix for the input data points, $\Phi_\mathbf{x}$ is a matrix of the expanded dimensionality (kernel algorithms) of the input data. The objective function is minimized with respect to the $A$ and $B$ matrices (they are considered to be regression matrices and decoupling the transformation $BA^t$ in two matrices allows the generalization of techniques like Canonical Correlation Analysis). The rank constraint is set to promote robustness of the solution towards a rank deficient $\Phi_\mathbf{x}$ matrix.

Hessian eigenmaps [Donoho2003] work with the Hessian of the graph instead of its Laplacian. By doing so, they extend ISOMAP and Laplacian eigenmaps, and they remove the need to map the input data onto a convex subset of $\mathbb{R}^m$.

Locally linear embedding (LLE) [Roweis2000, Saul2003] is another technique used to learn manifolds close to the data and project them onto it. For each observation $\mathbf{x}_n$ we look for the K-nearest neighbors (noted as $\mathbf{x}_{n'}$) and produce a set of weights for its approximation minimizing $\left\| \mathbf{x}_n - \sum_{n'} \beta_{nn'} \mathbf{x}_{n'} \right\|^2$ (see Fig. 18). This optimization is performed simultaneously for all observations, i.e. $J_{LLE} = \sum_n \left\| \mathbf{x}_n - \sum_{n'} \beta_{nn'} \mathbf{x}_{n'} \right\|^2 = \sum_n \left\| \mathbf{x}_n - X\boldsymbol{\beta}_n \right\|^2$, where $\boldsymbol{\beta}_n$ is a weight vector to be determined and whose value is non-zero only for the neighbors of $\mathbf{x}_n$. We can write this even more compactly by stacking all weight vectors as columns of a matrix $\boldsymbol{\beta}$, then $J_{LLE} = \left\| X - X\boldsymbol{\beta} \right\|_F^2$. The optimization is constrained so that the sum of $\boldsymbol{\beta}_n$ is 1 for all $n$. Once the weights have been determined, we look for lower dimension points, $\boldsymbol{\chi}_n$, such that $\sum_n \left\| \boldsymbol{\chi}_n - \sum_{n'} \beta_{nn'} \boldsymbol{\chi}_{n'} \right\|^2$, that is the new points have to be reconstructed from its neighbors in the same way (with the same weights) as the observations they represent. This latest problem is solved by solving an eigenvalue problem and also keeping the smallest eigenvalues. See Fig. 19 for a comparison of the results of LLE, Hessian LLE and ISOMAP in a particular case.

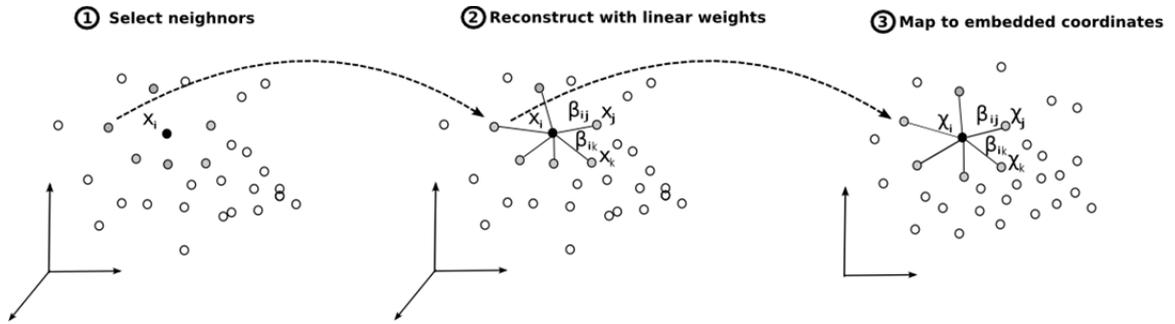

Figure 18. Schematic representation of the transformations involved in LTSA.

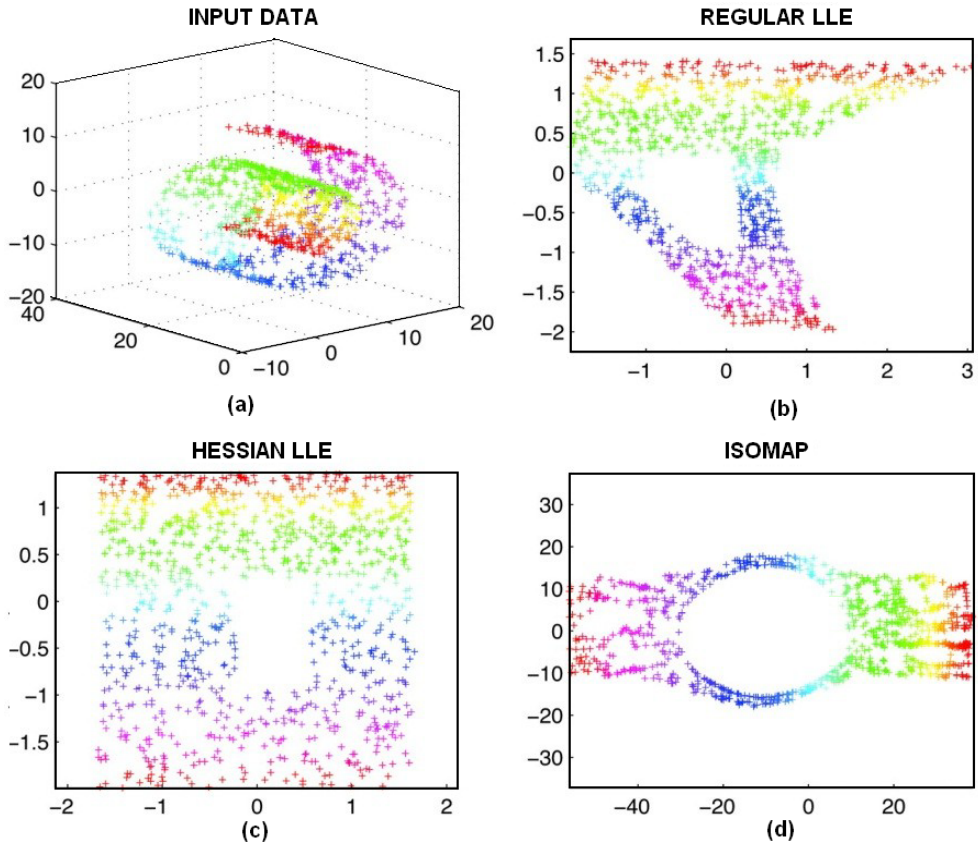

*Figure 19. Comparison of the results of LLE, Hessian LLE and ISOMAP for the Swiss roll input data.*

Latent Tangent Space Alignment (LTSA) [Zhang2004, Zhang2012] is another technique locally learning the manifold structure. As in LLE, we look for the local neighbors of a point. However, we now compute the local coordinates, $\chi'_n$, of all the input points in the PCA subspace associated to this neighborhood. Next, we need to align all local coordinates. For each input point we compute the reconstruction error from its coordinates in the different neighborhoods where it participated $J_{LTSA_n} = \sum_{n'} \|\chi_n - (c_{n'} + L_{n'}\chi'_n)\|^2$ ($c_{n'}$ is a vector and $L_{n'}$ is a matrix, both to be determined for all neighborhoods; we may think of them as translation and shape parameters that properly locate the different neighborhoods in a common geometrical framework). The objective function of LTSA is $J_{LTSA} = \sum_{n=1}^{N} J_{LTSA_n}$ that has to be optimized with respect to $\chi_n$, $c_{n'}$ and $L_{n'}$ (see Fig. 19).

One of the problems of the previous techniques (ISOMAP, Laplacian eigenmaps, Locally Linear Embedding, and Latent Tangent Space Alignment) is that they are only defined in a neighborhood of the training data, and they normally extrapolate very poorly. One of the reasons is because the mapping is not explicit, but implicit. Locality Preserving Projections (LPP) [He2004] tries to tackle this issue by constraining the projections to be a linear projection of the input vectors, $\chi_n = A^t \mathbf{x}_n$. The goal function is the same as in Laplacian eigenmaps. The $A$ matrix is of size $m \times M$ and it is the parameter with respect to which the $J_{LE}$ objective function is minimized. An orthogonal version has been proposed [Kokiopoulo2007], with the constraint $A^t A = I$. A Kernel version of LPP also exists [He2004]. As in all kernel methods, the idea is to map the input vectors $\mathbf{x}_n$ onto a higher dimensional space with a nonlinear function, so that the linear constraint imposed by $\chi_n = A^t \mathbf{x}_n$ becomes a nonlinear projection. A variation of this technique is called Neighborhood Preserving Embedding (NPE) [He2005] where $J_{NPE} = \chi_i^t M \chi_j$ being $M = (I-G)^t(I-G)$. This matrix approximates the squared Laplacian of the weight graph $G$ and it offers numerical advantages over the LPP algorithm. Orthogonal Neighborhood Preserving Projections (ONPP) [Kokiopoulou2007] extends the linear projection idea, $\chi_n = A^t \mathbf{x}_n$, to the LLE algorithm.

The idea of constructing linear projections for the dimensionality reduction can be performed locally, instead of globally (as in LPP, OLPP, NPE and ONPP). This has been proposed by [Wang2011]. The manifold is divided in areas in which it can be well approximated by a linear subspace (a manifold is locally similar to a linear subspace if the geodesic distance between points is similar to the Euclidean distance among those points). The division is a disjoint partition of the input data points $\mathbf{x}_n$ such that the number of parts is minimized and each local linear subspace is as large as possible. Within each partition a linear PCA model is adjusted. Finally, all models are aligned following an alignment procedure similar to that of LTSA.

# 5. Trends and Conclusions

We have analyzed the number of citations that the most relevant papers in each section have received in the last decade (2003-2012). In Table I we show the number of citations summarized by large areas as well as their share (%) for the different years. At the sight of this table we can draw several conclusions:

- The interest in the field has grown by a factor 3 in the last decade as shown by the absolute number of citations.
- By far, the most applied techniques are those based on the search of components in its different brands (ICA, PCA, FA, MDS, …), although the tendency in the last decade is to loose importance in favor of those techniques using projections (especially, projections onto manifolds) or dictionaries. This is a response to the non-linear nature of experimental data in most fields.
- Dimensionality reduction techniques based on projections and dictionaries are growing very fast in the last decade: both, in the number of new methods and in the application of those methods to real problems.
- Interestingly, old methods based on vector quantization keep nearly a constant market share meaning that they are very well suited to a specific kind of problems. However, those methods that tried to preserve the input data topology while doing the vector quantization have lost impact, mostly because of the appearance of new methods capable of analyzing manifolds.

We can further subdivide these large areas into smaller subareas. Table II shows the subareas sorted by total number of citations. After analyzing this table we draw the following conclusions:

- The analysis on manifolds is the clear winner of the decade. The reason is its ability to analyze non-linearities and its capability of adapting to the local structure of the data. Among the different techniques, ISOMAP, Locally Linear Embedding, and Laplacian Eigenmaps are the most successful. This increase has been at the cost of the non-linear PCA versions (principal curves, principal surfaces and principal manifolds) and the Self-Organizing Maps since the new techniques can explore non-linear relationships in a much richer way.
- PCA in its different versions (standard PCA, robust PCA, sparse PCA, kernel PCA, …) is still one of the preferred techniques due to its simplicity and intuitiveness. The increase in the use of PCA contrasts with the decrease in the use of Factor Analysis, which is more constrained in its modeling capabilities.
- Independent Component Analysis reached its boom in the middle 2000's, but now it is declining. Probably, it will remain at a niche of applications related to signal processing for which it is particularly well suited. But it might not stand as a general purpose technique. It is possible that this decrease also responds to a diversification of the techniques falling under the umbrella of ICA.
- Non-negative Matrix Factorization has experienced an important raise, probably because of its ability of producing more interpretable bases and because they are well suited to many situations in which the sum of positive factors is the natural way of modeling the problem.

- The rest of the techniques have kept their market share. This is most likely explained by the fact that they have their own niche of applications, which they are very well suited to.

Overall, we can say that dimensionality reduction techniques are being applied in many scientific areas ranging from biomedical research to text mining and computer science. In this review we have covered different families of methodologies; each of them based on different criteria but all chasing the same goal: reduce the complexity of the data structure while at the same time delivering a more understandable representation of the same information. The field is still very active and ever more powerful methods are continuously appearing providing an excellent application test bed for applied mathematicians.

## Acknowledgements

This work was supported by the Spanish Minister of Science and Innovation (BIO2010-17527) and Madrid government grant (P2010/BMD-2305). C.O.S. Sorzano was also supported by the Ramón y Cajal program.